%% file: anonymous-submission-latex-2026.tex
\documentclass[letterpaper]{article} 
\usepackage[]{aaai2026}  
\usepackage{times}  
\usepackage{helvet}  
\usepackage{courier}  
\usepackage[hyphens]{url}  
\usepackage{graphicx} 
\usepackage{natbib}  
\usepackage{caption} 
\usepackage{algorithm}
\usepackage{algorithmic}

\usepackage{newfloat}
\usepackage{listings}
\DeclareCaptionStyle{ruled}{labelfont=normalfont,labelsep=colon,strut=off} 
\floatstyle{ruled}
\newfloat{listing}{tb}{lst}{}
\floatname{listing}{Listing}

\usepackage{microtype}
\usepackage{url}
\usepackage{booktabs}

\usepackage{lineno}
\usepackage[utf8]{inputenc} 
\usepackage[T1]{fontenc}    
\usepackage{url}            
\usepackage{booktabs}       
\usepackage{amsfonts}       
\usepackage{nicefrac}       
\usepackage{microtype}      
\usepackage{xcolor}         
\usepackage{graphicx}
\usepackage{amsmath}
\usepackage{tikz}
\usepackage{pifont}
\usepackage[symbol]{footmisc}
\usepackage{enumitem}
\usepackage{listings}
\usepackage{xcolor}
\usepackage{tcolorbox}
\usepackage{caption}
\usepackage{capt-of,etoolbox}
\usepackage{subcaption}
\usepackage{caption}
\usepackage{multirow}
\usepackage{tikz}
\usepackage{cleveref}
\newcommand*\circled[1]{\tikz[baseline=(char.base)]{
            \node[shape=circle,draw,inner sep=0.5pt] (char) {#1};}}

\newcommand{\cmark}{\text{\ding{51}}}
\newcommand{\xmark}{\text{\ding{55}}}

\title{ENTER: Event Based Interpretable Reasoning for VideoQA}

\author{Hammad Ayyubi\thanks{Equal Contribution}\; $^\spadesuit$ 
\quad Junzhang Liu\footnotemark[1]\; $^\spadesuit$ \quad Ali Asgarov$^\dagger$ \quad Zaber Hakim$^\dagger$ \quad Najibul Sarker$^\dagger$ \\[4pt]
 Zhecan Wang$^\spadesuit$ \quad Chia-Wei Tang$^\dagger$  \quad Hani Alomari$^\dagger$  \quad Md.~Atabuzzaman$^\dagger$ \quad  Xudong Lin$^\spadesuit$ \\[4pt]
 Naveen Reddy Dyava$^\spadesuit$
 \quad Shih-Fu Chang$^\spadesuit$  \quad Chris Thomas$^\dagger$ \\[4pt]
$^{\spadesuit}$Columbia University \quad
$^{\dagger}$Virginia Tech
\\
 {\tt\small
   hayyubi@cs.columbia.edu, jl6262@columbia.edu
 }
}

\usepackage{bibentry}

\begin{document}

\maketitle


\input{sec/0_abstract}

\input{sec/1_intro}
\input{sec/2_related_works}

\input{sec/3_method}

\input{sec/4_experiments}

\input{sec/5_conclusion}
\bibliography{aaai2026}
\input{supp}

\end{document}

%% file: sec/0_abstract.tex
\begin{abstract}
In this paper, we present ENTER, an interpretable Video Question Answering (VideoQA) system based on event graphs. Event graphs convert videos into graphical representations, where video events form the nodes and event-event relationships (temporal/causal/hierarchical) form the edges. This structured representation offers many benefits: 1) Interpretable VideoQA via generated code that parses event-graph; 2) Incorporation of contextual visual information in the reasoning process (code generation) via event graphs; 3) Robust VideoQA via Hierarchical Iterative Update of the event graphs. Existing interpretable VideoQA systems are often top-down, disregarding low-level visual information in the reasoning plan generation, and are brittle. While bottom-up approaches produce responses from visual data, they lack interpretability. Experimental results on NExT-QA, IntentQA, NExT-GQA, and STAR demonstrate that not only does our method outperform existing top-down approaches while obtaining competitive performance against bottom-up approaches, but more importantly, it offers superior interpretability and explainability in the reasoning process.
\end{abstract}

%% file: sec/1_intro.tex
\input{figures/teaser_figure}
\section{Introduction}
\label{sec:intro}
Video Question Answering (VideoQA) \citep{nextqa, activitynetqa} is 
a challenging task that requires recognizing objects, people, actions, and events, and understanding their relationships' evolution across time.
Traditional VideoQA models \citep{internvideo, song2024moviechatdensetokensparse} were trained end-to-end, and 
while these systems perform strongly, they suffer from a lack of interpretability in their decision process.
This lack of transparency undermines trust and reliability and limits their utility when understanding the model's decision-making process is crucial. 

To address this, modular reasoning approaches \citep{vipergpt, morevqa} break down the reasoning process into interpretable steps,  
making the decision-making process more transparent, reliable, and trustworthy. 
Such properties are vital for applications in high-stakes environments, real-world deployments where human oversight is essential, and improving individual reasoning or perception modules through explicit feedback.
Despite their promise, current interpretable approaches for VideoQA \cite{vipergpt, morevqa} follow a top-down approach by generating reasoning plans solely from the input question. 
While their 
modular structure improves interpretability, it fails to fully utilize the visual modality.
Ignoring this visual context during plan generation limits the reasoning system's ability to incorporate contextual clues that are critical for answering complex questions or navigating complex media assets such as long-range videos. 
Further, these systems are brittle: if the key visual clues are missing from the initial plan, the system is not able to correct the incomplete plan. 
In contrast, 
bottom-up approaches \cite{llovi, videoagente2e} 
directly use visual frames to produce their answer. 
While some methods \cite{wang2024videotreeadaptivetreebasedvideo} select frames based on the context of questions which allows these systems to be contextual and robust, their reasoning process is still opaque since they directly produce the answer from captions using Large Language Models (LLMs).



%

To bridge this gap, we propose ENTER, \textbf{E}vent Based I\textbf{nte}rpretable \textbf{R}easoning. ENTER combines the interpretability of modular reasoning with the contextual awareness and robustness of bottom-up systems by integrating visual information into the plan generation process.
ENTER uses a structured and intermediate representation of videos known as event graphs, as shown in \cref{fig:teaser_method_comp}.
In these graphs, nodes represent events and edges capture relationships between events, forming a rich structured representation of the events in the video.
The event graph, along with the question, is passed to a large language model (LLM) to generate modular, Python-based code that parses the graph.
This process allows the system to explicitly model relationships between events and incorporate them into the reasoning process.
The generated code is then executed to arrive at the final answer.

ENTER uses a hierarchical approach to enhance graph completeness for answering questions while considering operation cost to maximize efficiency.
Initially, ENTER generates a dense graph from available captions, as this is the least costly operation. 
If the initial graph lacks necessary detail, ENTER enhances the captions and regenerates a denser graph. 
Only if these steps are insufficient does ENTER introduce multimodal edges, which is a more resource-intensive operation. 
By doing so, ENTER makes efficient use of computational resources but also enhances the robustness of the graph to mitigate failure cases caused by incomplete graphs.



ENTER's syntactic, interpretable code and event graph provide full transparency into its decision-making process. Event graphs provide ENTER with a powerful representation that captures explicit relationships between events throughout an entire video, enabling it to address questions requiring long-range dependencies. 
The structured design of these graphs naturally embeds complex semantic relationships, such as causality and temporal order, enabling robust, interpretable reasoning. 
ENTER combines the clarity and modularity of interpretable models with the contextual awareness of end-to-end systems, offering users a balanced, transparent, and effective reasoning framework.

We demonstrate the effectiveness of ENTER through comparable or even better performance than state-of-the-art on the NExT-QA \cite{nextqa}, IntentQA \cite{li2023intentqa}, NExT-GQA, \cite{xiao2024itrustanswervisually}, and STAR \cite{wu2024starbenchmarksituatedreasoning} VideoQA datasets
and highlight the benefits of structured event graphs for video reasoning. 
Our qualitative results highlight the advantages of this representation and illustrate how event-based graphs contribute to more accurate and
interpretable reasoning. 
Finally, our approach allows more focused debugging and easily allows isolation of the root cause of the error from the representation, the executable reasoning, or the reasoner's use of the knowledge.



%% file: figures/teaser_figure.tex
\begin{figure}[t]
    \centering
    \includegraphics[width=0.8\linewidth]{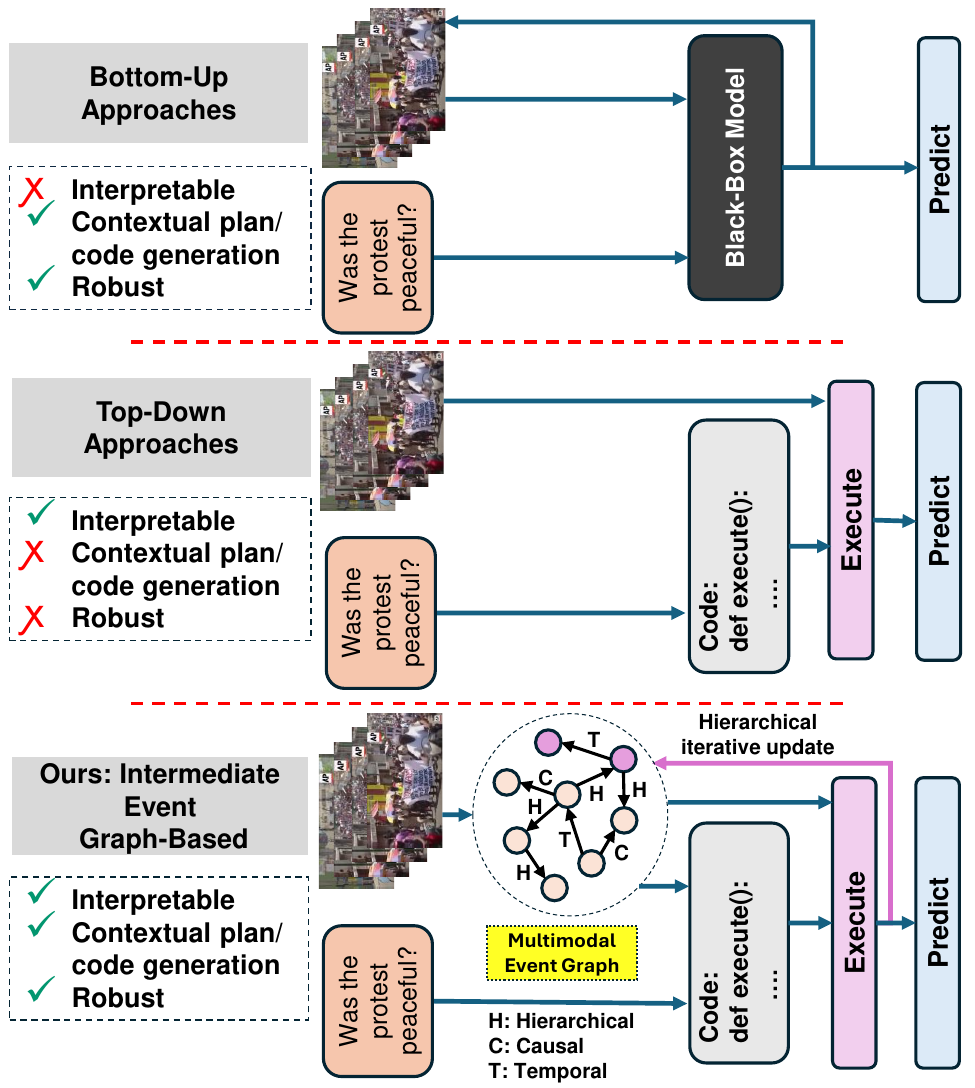}
    \caption{Comparison of our proposed method with existing 
    approaches for VideoQA. Bottom-up approaches directly process visual data but their reasoning is not transparent. In contrast, Top-down methods are interpretable but fail to use visual information in the plan or code generation piece.
    In contrast, our intermediate multimodal graph-based method combines the strengths of both: interpretability + use of visual information for event graph generation, making our predictions more transparent and context-aware.} 
    \label{fig:teaser_method_comp}
    \vspace{-0.4cm}
\end{figure}

%% file: sec/2_related_works.tex
\section{Related Works}

\subsection{Top-Down vs. Bottom-Up in Video QA}
Existing interpretable VideoQA methods \citep{mmreact, videoagenttopdown, contphy, vipergpt, morevqa} often take a top-down approach generating reasoning steps in response to a specific question. Typically, these steps are structured as programs that use APIs to interact with the video \cite{vipergpt, morevqa, subramanian2023modularvisualquestionanswering, visprog} or as chains of thought grounded in visual cues \cite{fei2024videoofthought}. However, because reasoning steps are generated independently of the visual data, 
their ability to fully understand video content is constrained. 
In contrast, bottom-up approaches \citep{lin2021vx2text,internvideo, ashutosh2023hiervllearninghierarchicalvideolanguage, islam2024videorecaprecursivecaptioning, videollama, videoagente2e,  llovi, videochat, videochatgpt, wang2024videotreeadaptivetreebasedvideo} jointly process both the question and video, allowing for reasoning that integrates richer visual information. For example, methods like LLoVi \cite{llovi} employ extensive captioning, while VideoAgent \cite{videoagente2e} iteratively selects keyframes. While these models can incorporate more detailed visual data, they often lack interpretability, typically providing direct answers without explicit reasoning paths. Both top-down and bottom-up approaches face a trade-off between interpretability and the depth of visual information used. To bridge this gap, recent work \citep{lin2023towards, hair, openpvsg, chen2024groundedmultihopvideoqalongform, videoagente2e, lin2024training} has explored end-to-end methods that aim to integrate visual data while maintaining interpretability. However, these methods still struggle to fully address complex queries, often providing limited reasoning steps or relying on simple visual or commonsense cues.

\begin{figure*}[t]
    \vspace{-0.5cm}
    \centering
    \includegraphics[width=\linewidth]{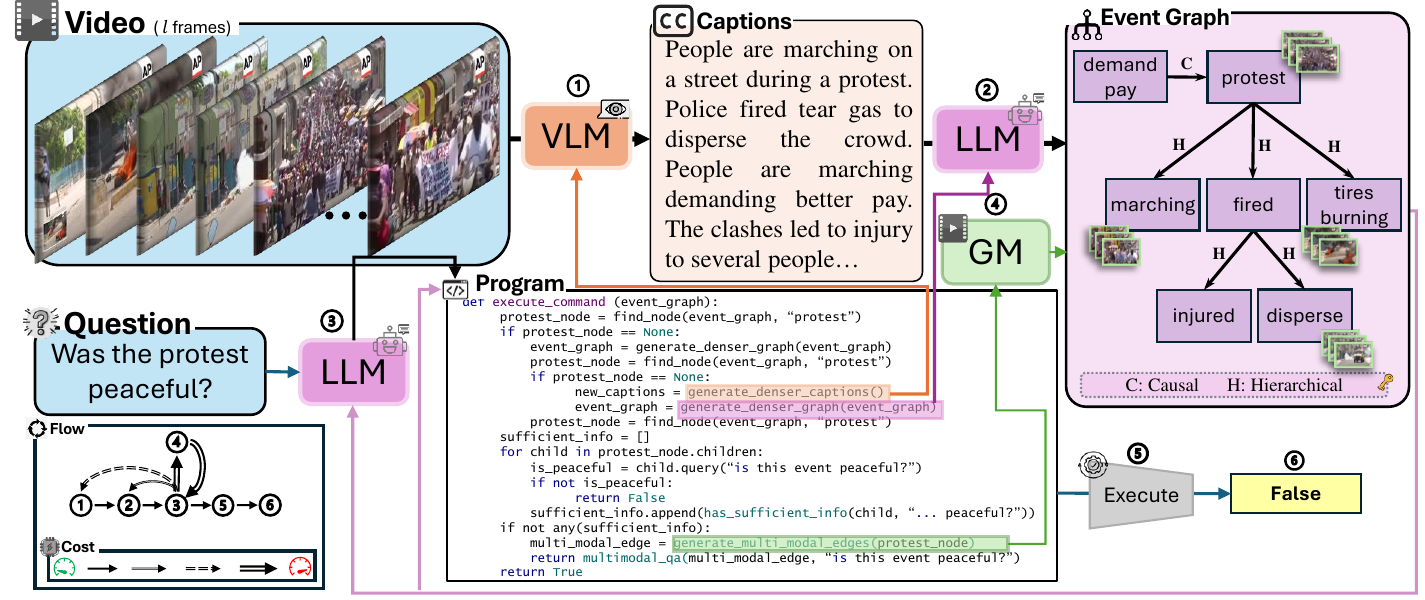}
    \caption{Overall pipeline of our method. Video is first converted to event-graph via captioning as the intermediate step. Next, python code is generated, contextualized by event-graph, to parse the graph. Further, any missing information in graph is added via iterative hierarchical update. Bottom-left flow chart depicts the flow of our method, along with the relative cost of each operation.}
    \label{fig:method}
    \vspace{-0.5cm}
\end{figure*}

\subsection{Contextual Plan Generation}

Our work is inspired by Neural Module Networks \cite{andreas2017neuralmodulenetworks}, which suggest that complex vision tasks are inherently compositional and should be divided into atomic perceptual units. The recent methods, CodeVQA \cite{subramanian2023modularvisualquestionanswering}, ViperGPT \cite{vipergpt}, and VisProg \cite{visprog} achieve accuracy comparable to end-to-end systems by leveraging GPT-3's \cite{brown2020languagemodelsfewshotlearners} and GPT-4's \cite{openai2024gpt4technicalreport} ability to generate code that acts as a plan for answering visual questions through reasoning steps.
This approach enables the use of logical operators to manipulate the outputs of visual models, however, they generate the entire reasoning plan in one pass without access to evolving visual context, limiting their adaptability. 
MoReVQA \cite{morevqa} 
introduces multi-stage plan generation but does not integrate visual context throughout its reasoning process. 
In contrast, our method can generate and adaptively refine the reasoning plan in response to evolving visual details from complex dynamic scenes in videos.

\subsection{Iterative Robustness in VideoQA}
Many existing VideoQA methods \citep{ko2023largelanguagemodelstemporal, morevqa, choudhury2023zeroshotvideoquestionanswering, mogrovejo-solorio-2024-question, pan2023retrievingtoanswerzeroshotvideoquestion, shen2024longvuspatiotemporaladaptivecompression, han2024freevideollmpromptguidedvisual, wu2024freevaofflinemllmtrainingfree, xu2024slowfastllavastrongtrainingfreebaseline, zhang2024simplellmframeworklongrange, wang2024stairspatialtemporalreasoningauditable} focus on question-answering without incorporating a comprehensive error analysis mechanism. 
This 
leads to models that are less robust 
when given complex or ambiguous queries. To solve this, TraveLER \cite{shang2024travelermodularmultilmmagent} iteratively refines its understanding through locating, evaluating, and replanning agents. 
VidF4 \cite{liang2024endtoendvideoquestionanswering} introduces frame-scoring mechanisms for question relevance and inter-frame similarity.
VideoInsta \cite{ liao-etal-2024-videoinsta} applies a self-reflective reasoning approach to enhance information sufficiency, confidence, and temporal balance in VideoQA.
VideoTree \cite{wang2024videotreeadaptivetreebasedvideo} iteratively refines keyframe selection based on query relevance through a hierarchical, tree-based structure. 
ENTER's self-correction strategy helps ensure a more thorough understanding of the video content when needed which improves the robustness and performance of the VideoQA system.

%% file: sec/3_method.tex
\section{Method}
\label{sec:method}
We propose a modular approach for VideoQA that represents videos with event graphs and generates code to parse the graph interpretably.
Starting with a detailed video description, we construct the event graph (\Cref{sec:event_graph_gen}), which, along with the input question, directs the generation of a Python-based reasoning plan (\Cref{subsec:code_gen}). This code is then executed to answer the question or, if details are missing, performs self-correction by iteratively retrieving necessary information (\Cref{sec:iterative_graph_update}) for improved robustness.

\subsection{Event Graph Generation}
\label{sec:event_graph_gen}
To generate event graphs from videos, we first convert the video to captions using Vision Language Model (VLM) and then extract event graphs from captions using LLMs. 
The two stages leverage the strength of VLM and LLM for each of their respective tasks in the process. In comparison, a single-stage process to convert videos to event graphs suffers from noisy predictions since VLMs were not inherently trained on graph generation tasks.

\subsubsection{Caption Generation from Video}
\label{sec:caption_gen}
To build our event graph, we generate structured descriptions of video events. Standard captioning methods, like frame-based captioning \citep{zhang2024simplellmframeworklongrange}, often lose entity coherence and miss subtle event relationships, limiting their effectiveness for detailed graph construction. Although recent work \citep{raajesh2024micapunifiedmodelidentityaware} enhances coreference for people, it lacks depth in capturing complex relationships for generic objects.


Our approach (shown as \circled{1} in \Cref{fig:method}) emphasizes preserving event relationships, providing detailed descriptions, and maintaining consistent entity coreference. We prompt the model to use relational keywords like “after,” “due to,” or “meanwhile” to capture temporal, causal, and hierarchical links. Entities are described specifically (e.g., “white Honda Civic”) and consistently referenced (e.g., “black car with broken window”) throughout. This structured captioning supports event graph generation that links entities and relationships across the video, essential for robust reasoning.

\subsubsection{Graph Generation}
To build the event graph (shown as \circled{2} in \Cref{fig:method}), we extract events and their relationships from the caption, where each node represents an event, and directed edges capture intra-event semantic relationships: temporal, causal, and hierarchical. Temporal edges link events occurring sequentially, causal edges connect events with cause-effect relationships, and hierarchical edges denote event containment.

Each node includes arguments that represent participants or objects involved in the event (e.g., in ``A man driving a car,'' the event ``Drive'' includes ``actor'' as ``man'' and ``vehicle'' as ``car''). To preserve coreference, we prompt the model for consistent naming of arguments and events across the graph, as discussed in \Cref{sec:caption_gen}. Additionally, the model is prompted to identify all relevant event relationships to produce a comprehensive, cohesive graph that captures the video’s full semantic narrative. 

To ensure the information recorded in the graphs are correct and sufficient to answer the question, we sampled 341 event nodes and 853 edges. We manually verified their correctness, i.e. the event/relation described is indeed present in the video. We also sample 50 graphs to verify if the graph itself is sufficient to answer the question. As shown in \Cref{table:check_event_graph}, the generated graph nodes have a precision of 96.5\% and edges are 87.7\%, meaning the event graphs are precise descriptions of the videos. The sufficiency of the event graphs is 92.0\%, meaning most questions are sufficiently answerable given the event graphs.
\input{tables/event_graph_check}

\subsection{Code Generation and Execution}\label{subsec:code_gen}
To answer questions, we generate Python code that utilizes both the event graph and the textual question. Inspired by ViperGPT’s API with image patches \citep{vipergpt}, we designed an API with \texttt{Event} and \texttt{EventGraph} classes to navigate and query the graph. For example, \texttt{EventGraph} includes a \texttt{get\_child} method to retrieve hierarchically related nodes for a given event. The code generator receives an abstract API definition with function signatures and examples. This API and prompt guide the model to produce Python code focused on task-specific logic, executed via Python’s interpreter using basic control structures like \texttt{if/else} statements and loops. This abstraction improves efficiency by focusing the model on high-level reasoning.

To derive the final answer, we extract the relevant event and arguments from the question, locate the corresponding node in the graph, and traverse its subgraph to collect the necessary details. Unlike other VideoQA methods that may overlook visual features or process video data inefficiently, our approach uses the event graph’s compact, text-based representation to integrate visual context efficiently in the code generation process. 
Code generation and execution correspond to steps \circled{3} and \circled{5} in \Cref{fig:method}. In case of multiple choice question, a reasoner (LLM) call is made via \texttt{select\_answer(question, possible\_answers, info)}.

\input{tables/main_result}

\subsection{Iterative Hierarchical Graph Update}
\label{sec:iterative_graph_update}
In cases where the initial event graph lacks the details necessary to answer the question, we perform an Iterative Hierarchical Graph Update process to address information gaps. 
Missing information can result from limitations in the captioning process, graph generation or mismatches between video and text information (modality gap). Unlike existing modular methods that risk undetected gaps or hallucinated results, our graph-based approach identifies and addresses these gaps through a structured, self-correcting update process.
The update mechanism applies a three-stage, layered approach. Starting with low-cost graph consolidation, it progressively adds detail by refining captions and, if needed, incorporating multimodal data. 
The hierarchical refinement, described below, iteratively refines the event graph with additional layers of information only as needed which improves the system's ability to accurately answer questions, while remaining resource-efficient.



\noindent\textbf{Denser Graph Generation.}
If the missing information exists in the captions but is not captured in the graph, we generate a denser graph by reprocessing the captions with a focus on the missing details. 
We implement \texttt{generate\_denser\_graph(graph, caption, request)}, which reinvokes the graph generator using the current graph, captions, and a request to include the missing nodes or relations. This stage corresponds to the \circled{3} to \circled{2} \Cref{fig:method}.

\noindent\textbf{Denser Caption Generation.}
When necessary events or relations are absent from the captions themselves, we generate denser captions by prompting the captioner to include the missing information. The function \texttt{generate\_denser\_caption(caption, video, request)} reprocesses the video to extract these details. The new information is then integrated into the graph using \texttt{generate\_denser\_graph}. This stage is represented by the \circled{3} to \circled{1} loop in \Cref{fig:method}. We repeat this procedure a fixed number of times or until the relevant events and relations are included before proceeding to the next step.

\noindent\textbf{Multimodal Information Incorporation.}
In cases where the denser captions and graph remain insufficient to answer the question, we enhance the graph using multimodal information. Uncertainty in the model's answers can be detected using confidence assessments \citep{you2023idealgptiterativelydecomposingvision}. 
Upon detecting uncertainty, we invoke \texttt{get\_new\_info(question, request, video)}, prompting the captioner to revisit the video and retrieve the required information. 
If gaps remain and are detected by \texttt{has\_sufficient\_info}, as the last fail-safe, we enhance the graph nodes with relevant video clips to create a multimodal representation. 
Specifically, \texttt{generate\_multi\_modal\_edges(subgraph)}, iterates through the nodes of the subgraph (Section \ref{subsec:code_gen}) and retrieves node-relevant video clips using Gemini 1.5 Flash as the text-to-video retrieval model.
The enhanced multimodal subgraph is then passed to a vision-language model to answer the question. This stage corresponds to the \circled{4} to \circled{3} loop in \Cref{fig:method}.

We list all the prompts used in our method in \Cref{sec:prompts}, and we will release our code on acceptance.

%% file: tables/event_graph_check.tex
\begin{table}[t]\small \centering
\vspace{-0.3cm}
\centering
\resizebox{0.3\textwidth}{!}{\begin{tabular}{l|l|l}
        \toprule
        Prec. Nodes& Prec. Edges& Sufficiency \\
        \midrule
        96.5&87.7&92.0\\
        \bottomrule
    \end{tabular}}
    \caption{Precision of nodes and edges measure the percentage of nodes/edges present in the graph that are indeed in the video, i.e. the described events/relations are not hallucinated, fabricated, or incorrect. Sufficiency measures the percentage of questions that are sufficiently answerable given the graph.}
    \label{table:check_event_graph}
    \vspace{-0.3cm}
\end{table}

%% file: tables/main_result.tex
\begin{table*} [h!]\small
\vspace{-0.3cm}
    \centering
    
\resizebox{0.9\textwidth}{!}{
    \begin{tabular}{c|cccc|cccc|ccccc}
        \toprule
        \multirow{2}{*}{Method} & \multicolumn{4}{c|}{NExT-QA} & \multicolumn{4}{c|}{IntentQA}  & \multicolumn{5}{c}{STAR}\\
        &All &Cau.&Tem.&Des. & All & CW & CH & TP\&TN &Mean & Int&Seq&Pre&Fea\\
        \midrule
         \textcolor{gray}{\textit{End-to-End}}&&&&&&&&&\\
        \textcolor{gray}{VideoChat2 HD  \citep{li2024mvbenchcomprehensivemultimodalvideo}}& \textcolor{gray}{61.7}& \textcolor{gray}{61.9}& \textcolor{gray}{57.4}& \textcolor{gray}{69.9} & \textcolor{gray}{-}& \textcolor{gray}{-}& \textcolor{gray}{-}& \textcolor{gray}{-}&\textcolor{gray}{-}&\textcolor{gray}{-}&\textcolor{gray}{-}&\textcolor{gray}{-}&\textcolor{gray}{-}\\
         \textcolor{gray}{SeViLA \citep{yu2023selfchainedimagelanguagemodelvideo}}&  \textcolor{gray}{63.6}& \textcolor{gray}{61.5}& \textcolor{gray}{61.3}& \textcolor{gray}{75.6}& \textcolor{gray}{60.9}& \textcolor{gray}{-}& \textcolor{gray}{-}& \textcolor{gray}{-}& \textcolor{gray}{44.6}&\textcolor{gray}{-}&\textcolor{gray}{-}&\textcolor{gray}{-}&\textcolor{gray}{-}\\
         \textcolor{gray}{Gemini 1.5 Flash}& \textcolor{gray}{70.4}& \textcolor{gray}{71.9}& \textcolor{gray}{64.1}& \textcolor{gray}{78.1}& \textcolor{gray}{68.8}& \textcolor{gray}{70.8}& \textcolor{gray}{76.2}& \textcolor{gray}{59.2}& \textcolor{gray}{-}&\textcolor{gray}{-}&\textcolor{gray}{-}&\textcolor{gray}{-}&\textcolor{gray}{-}\\
        \midrule
         \textcolor{gray}{\textit{Bottom-up}}&&&&&&&&\\
         \textcolor{gray}{LLoVi \citep{zhang2024simplellmframeworklongrange}}& \textcolor{gray}{67.7}& \textcolor{gray}{69.5}& \textcolor{gray}{61.0}& \textcolor{gray}{75.6}& \textcolor{gray}{67.1}& \textcolor{gray}{-}& \textcolor{gray}{-}& \textcolor{gray}{-}&\textcolor{gray}{-}&\textcolor{gray}{-}&\textcolor{gray}{-}&\textcolor{gray}{-}&\textcolor{gray}{-}\\
         \textcolor{gray}{VideoTree \citep{wang2024videotreeadaptivetreebasedvideo}}&  \textcolor{gray}{75.6} &  \textcolor{gray}{76.5}& \textcolor{gray}{70.6}& \textcolor{gray}{83.9}& \textcolor{gray}{66.9}& \textcolor{gray}{-}& \textcolor{gray}{-}& \textcolor{gray}{-}& \textcolor{gray}{-}&\textcolor{gray}{-}&\textcolor{gray}{-}&\textcolor{gray}{-}&\textcolor{gray}{-}\\
         \textcolor{gray}{LVNet \citep{park2024framesusefulefficientstrategieslongform}} & \textcolor{gray}{72.9}& \textcolor{gray}{75.0}& \textcolor{gray}{65.5}& \textcolor{gray}{81.5}& \textcolor{gray}{71.7}& \textcolor{gray}{75.0}& \textcolor{gray}{74.4}& \textcolor{gray}{62.1}& \textcolor{gray}{-}&\textcolor{gray}{-}&\textcolor{gray}{-}&\textcolor{gray}{-}&\textcolor{gray}{-}\\
         
         \textcolor{gray}{Q-ViD \citep{zhang2024simplellmframeworklongrange}}& \textcolor{gray}{66.3}& \textcolor{gray}{67.6}& \textcolor{gray}{61.6}& \textcolor{gray}{72.2}&  \textcolor{gray}{-}& \textcolor{gray}{-}& \textcolor{gray}{-}& \textcolor{gray}{-}& \textcolor{gray}{45.7}&\textcolor{gray}{48.2}&\textcolor{gray}{47.2}&\textcolor{gray}{43.9}&\textcolor{gray}{43.4}\\
         
         \textcolor{gray}{VideoINSTA\citep{liao-etal-2024-videoinsta}} & \textcolor{gray}{72.3}& \textcolor{gray}{-}& \textcolor{gray}{-}& \textcolor{gray}{-}&  \textcolor{gray}{72.8}& \textcolor{gray}{-}& \textcolor{gray}{-}& \textcolor{gray}{-}&  \textcolor{gray}{-}&\textcolor{gray}{-}&\textcolor{gray}{-}&\textcolor{gray}{-}&\textcolor{gray}{-}\\
         
        \midrule
        \textit{Top-Down/Modular}&&&&&&&&&\\
        ProViQ \citep{choudhury2023zeroshotvideoquestionanswering}&63.8&{-}&{-}&{-}& {-}& {-}& {-}& {-}& -&-&-&-&-\\
        MoReVQA \citep{morevqa}&69.2&61.3&61.5&75.6& {-}& {-}& {-}& {-}& -&-&-&-&-\\
        TraveLER \citep{shang2024travelermodularmultilmmagent}&68.2&70.0&60.5&78.2& 65.2& 69.9& 64.7& 54.4& 44.9&-&-&-&-\\
        ViperGPT \citep{vipergpt}&60.0&-&-&-& 54.7& {-}& {-}& {-}& 41.6&-&-&-&-\\
        Ours &\textbf{75.1}& 77.9 &68.2&79.2&\textbf{71.5}&73.2& 79.1& 61.4 & \textbf{67.1}&63.9&69.1&67.2&69.1\\
        \bottomrule
    \end{tabular}}
    \caption{Our method on NExT-QA, IntentQA, and STAR compared with other zero-shot methods. Our method outperforms most of them while maintaining interpretability.}
    \label{table:next_qa_result}
    \vspace{-0.3cm}
\end{table*}

%% file: sec/4_experiments.tex
\section{Experiments}
\label{sec:exp}

To validate the effectiveness of our approach, we conduct a series of experiments on the NExT-QA \citep{xiao2021nextqanextphasequestionansweringexplaining}, IntentQA \citep{Li_2023_ICCV}, NExT-GQA, \cite{xiao2024itrustanswervisually}, and STAR \citep{wu2024starbenchmarksituatedreasoning} datasets. These experiments are designed to  provide our system's performance on multiple-choice question-answering tasks and comparisons against existing methods. In addition to performance metrics, we also focus on the interpretability of our approach. The following sections show our experimental setup and results.

\subsection{Experimental Setup}
Our system employs Gemini 1.5 Flash for captioning, graph generation, and code generation, with GPT-4 serving as the reasoner. Accuracy is used as the primary evaluation metric across all the multiple-choice VideoQA datasets.

\textbf{NExt-QA\citep{xiao2021nextqanextphasequestionansweringexplaining}:} This dataset contains 4996 multiple choice questions, corresponding to 570 videos. The average length of the video clips is 43 seconds. The questions are categorized into three types: Causal (Cau), Temporal (Tem), and Descriptive (Des). 

\textbf{IntentQA\citep{Li_2023_ICCV}:} This dataset specifically focuses on inference-based QA types, including Causal and Temporal questions. It includes two subtypes under Causal— Causal Why (CW) and Causal How (CH), and two under Temporal—Temporal Previous (TP) and Temporal Next (TN). The dataset contains 2,134 questions corresponding to 567 videos.

\textbf{NExT-GQA\cite{xiao2024itrustanswervisually}:} Built upon the existing videos of NExT-QA. It not only tests the model's raw accuracy, but also requires the model to ground its answer in the video by providing the evidence in the form of time stamps. The correctness of grounding are measured in intersection over union (IoU) and intersection over prediction (IoP). Only correctly grounded answers (i.e. IoP$\geq$0.5) are credited when calculating the GQA accuracy.

\textbf{STAR\citep{wu2024starbenchmarksituatedreasoning}:} This dataset focuses on the ability to perform logical reasoning in the form of multiple-choice question-answering on real-world videos. 
It offers two evaluation splits: an evaluation set of 7,098 video-question pairs with publicly available answers, and a test set of 7,377 video-question pairs with hidden answers. We report our results on the evaluation set following the work of papers such as TraveLER \citep{shang2024travelermodularmultilmmagent} and SeViLA \citep{yu2023selfchainedimagelanguagemodelvideo}. 

NExT-QA, IntentQA, and STAR are particularly suited for evaluation of our approach as they are event-centric and require strong causal and temporal reasoning. 

\input{tables/next_gqa}
\noindent\textbf{Baselines.} 
We evaluate our approach against three categories of baselines: \textit{End-to-End}, \textit{Bottom-Up}, and \textit{Top-Down/Modular}. \textit{End-to-End} models, such as large vision-language models like Gemini, are typically not interpretable and may be pre-trained on VideoQA datasets. Similarly, \textit{Bottom-Up} approaches such as VideoTree \cite{wang2024videotreeadaptivetreebasedvideo} also lack interpretability. As such, these models are not directly comparable to our method.
Our primary comparisons are with \textit{Top-Down/Modular} methods, which emphasize interpretability. The most direct comparison for our work is against prior modular approaches like MoReVQA \cite{morevqa}, which provide transparent reasoning processes similar to ours.

\subsection{Experiments Results}



\Cref{table:next_qa_result} summarizes our results. We make the following observations.

\noindent\textbf{Our approach achieves outperforms top-down approaches on NExT-QA.} 
Compared to the best modular approach, MoReVQA \citep{morevqa}, we achieve a 16.6\% improvement on causal questions and a 6.7\% improvement on temporal questions and achieve comparable accuracy to the SOTA approach.
For descriptive questions, we outperform all modular methods and achieve comparable accuracy to the best bottom-up approach, VideoTree \citep{wang2024videotreeadaptivetreebasedvideo}.

\noindent \textbf{Our performance on IntentQA is very close to the existing state-of-the-art method.} We achieve comparable performance against the best-performing Bottom-Up method. The accuracy of our method slightly lags behind VideoINSTA\citep{liao-etal-2024-videoinsta} by 1.3\% and LVNet\citep{park2024framesusefulefficientstrategieslongform} by 0.5\%. However, our method offers interpretability, with a small compromise in accuracy. No modular approaches report benchmark scores on the IntentQA dataset, so a direct comparison against those methods was not possible.

\noindent\textbf{We outperform all zero-shot and top-down approaches on STAR.}
Our method surpasses all top-down and zero-shot approaches on the STAR dataset, with more than 20\% improvement over modular approaches like ViperGPT \citep{vipergpt} and TraveLER \citep{shang2024travelermodularmultilmmagent}. 
This considerable improvement highlights that our approach in handling complex, real-world event-based reasoning tasks require a deeper understanding of the event relations in video content. Additional ablation experiment on STAR is presented in \Cref{sec:additional_exp}.

\noindent\textbf{We achieve SOTA result on NExT-GQA.} As shown in table \Cref{table:next_gqa}, our model outperforms current SOTA method by 7.2\% on GQA accuracy. For video grounding metrics, we achieved 5\% and 6\% increase on mIoP and IoP@0.5 than SOTA score, and 9.8\% and 11.4\% increase on mIoU and IoU@0.5. This indicates that our method not only has transparent and interruptible reasoning process, but can also achieve grounded question-answering.

\input{tables/ablation_reasoner_comparison1}
In sum, our method surpasses all prior modular approaches on the NExT-QA, IntentQA, and STAR datasets and achieves comparable or superior performance to bottom-up methods on event-centric datasets like NExT-QA and IntentQA. While bottom-up models perform better on tasks requiring detailed video understanding, our model offers a clear advantage in transparency in its decision-making process—something unachievable with non-interpretable models.

We provide experiments on additional datasets in \Cref{sec:long_video_exp}.

\subsection{Ablation}
\noindent\textbf{All of ENTER's design components make an important contribution.}
\Cref{table:ablation_components} demonstrates the effectiveness of each of our model components on NExT-QA. Denser graph and Denser captions update mechanism boost performance from 63.8\% to 66.3\% and 67.2\% respectively (row 2 and 3). Further, updating the graph iteratively until 4 loops pushes the performance to 71.5\% (row 6).
Finally, we see that adding multimodal edges to compensate for lost visual information further improves the performance to 75.1\% (last row).
We provide further quantitative and qualitative ablations on the hierarchical update mechanism in \Cref{sec:hierarchical_update_ablation}.
In addition, we constructed a simple baseline with the LLMs we used -- Gemini and GPT-4o -- to isolate the gains from ENTER's event graph/code-based parsing. Gemini captioned the video while GPT-4o acted as the reasoner. The baseline's performance -- 70.5\% -- lagged significantly behind ENTER's (75.1\%), demonstrating our method's contributions.
\input{tables/ablation_denser_caption_graph}


\noindent\textbf{Reliable performance across both open-source and proprietary reasoners.} To evaluate the robustness of our approach to reasoning models, in \Cref{table:reasoner_ablation} we show ablations on using open-source to proprietary LLMs as the reasoner, . There is a 2.5\% difference in performance between the Llama 3.1 \citep{dubey2024llama3herdmodels} and the GPT4. The low fluctuation shows our method remains robust across different sized LLMs. 

\noindent\textbf{ENTER stays consistent, while other models fluctuate across reasoner types.} We further ablate against other approaches while keeping an open-source LLM (Llama 3) as the reasoner model across all systems to ensure a fair comparison. As shown in \Cref{table:llama3_compare}, when using the same open-source reasoner, our method overwhelmingly outperforms LloVi \cite{llovi} and VideoINSTA \cite{liao-etal-2024-videoinsta} by 25.9\% and 14.2\% in NExT-QA, and 20.1\% and 16\% in IntentQA respectively. Our model comfortably outperforms other models when open-source LLMs are considered. Interestingly, our performance difference between open-source Llama 3 and closed-source GPT4 is only around 2.6\% and 2.5\% in NExT-QA and IntentQA respectively. Comparing to LLoVi \cite{llovi} and VideoINSTA \cite{liao-etal-2024-videoinsta}, the score difference is more than 21\% and 18\% for NExT-QA, and 14\% and 19\% for IntentQA. Existing methods show huge degradation in performance when using open-source reasoners, while our method excels. This emphasizes our method is less dependent on the choice of reasoner.

\input{figures/positive_example}
\input{tables/ablation_scene_graph}
\noindent\textbf{Comparison Against Scene Graphs}
The graphical nature of scene graphs draw natural comparisons against event-graphs.
Scene graphs capture fine-grained object-level details -- \textit{subject, object, and predicates}. A closer look at event graphs shows that they are a super set of scene graphs. The \textit{subject, object, and predicates} relationships are captured in events and their entity arguments (\textit{actor, object, item, place, etc}).
In addition, event graphs contain explicit high-level temporal, hierarchical, and causal event-event relationships, making them far more expressive. 
To prove this hypothesis, we generated scene graphs for 38 videos from the NExT-QA dataset and asked human annotators to judge whether the information contained in the given graph was sufficient to answer the associated question. Annotators reported that the scene graphs were sufficient in only 21\% of the cases. In contrast, our event graph representations achieved a sufficiency rate of 92\%, as shown in \Cref{table:event_graph_vs_scene_graph}. This difference highlights the superior expressiveness of event graphs for complex reasoning tasks.

\noindent\textbf{Accuracy and Interpretability of Intermediate Steps} We conducted a human study on 30 randomly selected videos from the NExT-QA validation set. Annotators were asked to evaluate (1) whether the reasoning steps correctly led to the final answer, and (2) whether the reasoning steps were interpretable. Results show that in 66.67\% of cases, the reasoning steps aligned with the correct answer, and in 86.67\% of cases, they were judged interpretable. Considering ENTER achieves 75.1\% accuracy on NExT-QA, this implies that when ENTER answers correctly, its reasoning is valid in 88.89\% of those cases. These findings suggest that our method not only produces accurate answers, but also supports human-understandable reasoning in most cases.

\section{Qualitative Analysis}

We present two qualitative examples in \Cref{fig:samples_positive}. In the first example (left figure), the task is to count the total number of people in the video. The model iterates through event nodes, identifying human-related arguments, which are then counted by the final LLM reasoner using the \texttt{simple\_query} function. The second example (right figure) involves a more complex scenario. The question asks about the dog's action before being petted, which is missed by the captioner. The graph generator linked the man sitting on the chair as the event before petting. As this information dealt with the action of the man rather than the dog, it was insufficient. Consequently, \texttt{generate\_multimodal\_edges} retrieved the relevant clips. The dog's action, "Chasing its tail," occurring simultaneously with "Sitting," was retrieved, providing the correct answer. The examples depict the reasoning employed by our approach, showing its interpretability.
Additional examples in \Cref{sec:more_qualitative_samples}.

\subsection{Error Analysis}
The interpretability offered by our approach enables the diagnosis of different types of errors that our model makes and traces the source of the error. 
We categorized the errors into three types primarily.

    \noindent\textbf{Error Type: Insufficient Knowledge} occurs when the model lacks the necessary knowledge to properly interpret the scene. 

    \noindent\textbf{Error Type: Inconsistent Referencing} occurs when the model misreferences the same object or entity as two distinct ones due to inconsistent labeling. When the reasoning requires references to specific entities (i.e. counting, multiple actions of the same entity etc.), the information retrieval might be confused when inconsistent referencing happens. 
    
    \noindent\textbf{Error Type: Missed Information} occurs when the model misses key details needed to answer correctly. 

Please refer to \Cref{sec:error_analysis} for detailed analysis and qualitative examples.

%% file: tables/next_gqa.tex
\begin{table} [t]\small
\vspace{-1em}\fontsize{8}{9}\selectfont\setlength{\tabcolsep}{3pt}

\resizebox{0.45\textwidth}{!}{
    \begin{tabular}{l|llllll}
        \toprule
         Model&Raw QA &ACC@GQA&mIoP&IoP@0.5&mIoU&TIoU@0.5\\
        \midrule
        LLoVi& 66.8&24.3&37.3&36.9&20.0&15.3\\
        DeVi\cite{qin2024questionansweringdensevideoevents}&71.6&28.0&39.3&37.9&22.3&17.4\\
        MoReVQA&-&31.0&44.1&43.0& 20.7& 18.3\\
        \midrule
        Ours&\textbf{75.49}&\textbf{38.2}&\textbf{49.1}&\textbf{49.0}&\textbf{32.1}&\textbf{28.8}\\
        
        \bottomrule
    \end{tabular}}
    \caption{Grounded VideoQA results on NExT-GQA. Our method significantly outperforms other methods on all metrics. }
    \label{table:next_gqa}
    \vspace{-0.5cm}
\end{table}

%% file: tables/ablation_reasoner_comparison1.tex
\begin{table} [t]\small
    \begin{subtable}{.2\linewidth}
    \fontsize{7.5}{9}\selectfont
    \setlength{\tabcolsep}{3pt}
    
    \resizebox{0.98\textwidth}{!}{\begin{tabular}{l|l}
        \toprule
        Model & Acc. \\
        \midrule
        Llama 3&72.5\\
        Llama 3.1&72.6\\
        GPT4&\textbf{75.1}\\
        
        \bottomrule
    \end{tabular}}
    \caption{Reasoner ablation on NExT-QA.}
    \label{table:reasoner_ablation}
    \end{subtable}%
    \begin{subtable}{.8\linewidth}
    \label{table:multimodal_context}
    \setlength{\tabcolsep}{3pt}
    \fontsize{7.5}{9}\selectfont\centering
    \setlength{\tabcolsep}{3pt}
    \resizebox{0.93\textwidth}{!}{\begin{tabular}{l|l|l|l}  
        \toprule
        Model & System&NExT-QA & IntentQA \\
        \midrule
        \multirow{3}{*}{Llama 3}&LLoVi \cite{llovi}&46.6 \textcolor{gray}{$\downarrow $21.1}&48.9 \textcolor{gray}{$\downarrow $18.2}\\
        &VideoINSTA \cite{liao-etal-2024-videoinsta}&58.3 \textcolor{gray}{$\downarrow $14.0}&53.0 \textcolor{gray}{$\downarrow $19.8}\\
        &Ours&\textbf{72.5} \textcolor{gray}{$\downarrow $\hspace{0.6em}2.6}&\textbf{69.0} \textcolor{gray}{$\downarrow $\hspace{0.6em}2.5}\\        
        \bottomrule
    \end{tabular}}
    \captionsetup{justification=raggedleft}
    \caption{Comparison with other methods using open-source LLM reasoner. Gray values indicate performance difference from the GPT4 reasoner.}
    \label{table:llama3_compare}
    \end{subtable} 
    \caption{(a) Performance comparison of our methodology using different LLMs as the reasoner. (b) We compare our method with others using an open-source language model as the reasoner. Here, the performance difference from using a proprietary language model as the reasoner is shown in gray. }
    \vspace{-0.6cm}
\end{table}

%% file: tables/ablation_denser_caption_graph.tex
\begin{table} [t]\small \centering
\vspace{-1.4em}
\centering
    \fontsize{8}{9}\selectfont
    \begin{tabular}{ccccc}
        \toprule
        Den. Gph&Den. Cap.&Gph Type&\# Iter &Acc. \\
        \midrule
        \xmark &\xmark&Text &1&63.8\\
        \cmark &\xmark &Text &1&66.3\\
        \cmark &\cmark&Text&1&67.2\\
        \midrule
        \cmark &\cmark&Text&2&70.8\\
        \cmark &\cmark&Text&3&71.1\\
        \cmark &\cmark&Text&4&71.5\\
        \cmark &\cmark&Text&5&71.5\\
        \midrule
        \cmark &\cmark&Video&4&75.0\\
        \cmark &\cmark&MM (parent node)&4&74.3\\
        \cmark &\cmark&MM (subgraph)&4&\textbf{75.1}\\
        \bottomrule
    \end{tabular}
    \caption{Ablation of the denser graph generator, denser caption generator, multimodal graph generator, and iterative answer retrying mechanisms. Here MM denotes multimodal. Activating the denser graph (Den. Gph) and caption (Den. Cap.), with four iterations (\#Iter) of retrying, along with multimodal subgraph gives the best result.}
    \label{table:ablation_components}
\vspace{-0.3cm}
\end{table}

%% file: figures/positive_example.tex
\begin{figure*}[t]
    \includegraphics[width=\textwidth,keepaspectratio]{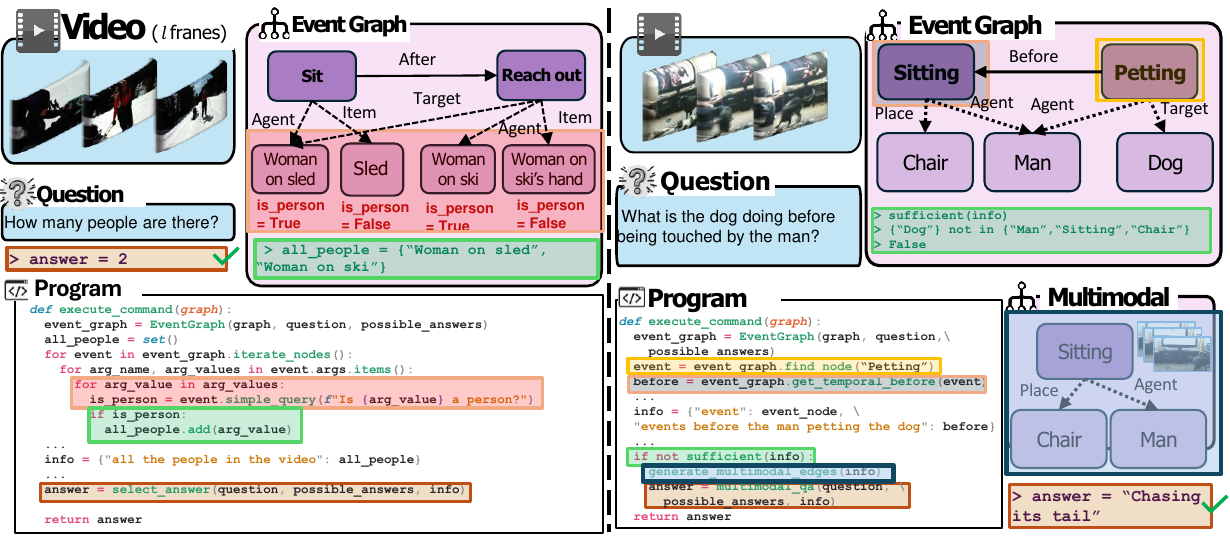}
    \caption{Qualitative examples of ENTER on NExT-QA. The output shows interpretability while maintaining correctness.}
    \label{fig:samples_positive}
    \vspace{-0.5cm}
\end{figure*}

%% file: tables/ablation_scene_graph.tex
\begin{table}[t]\small \centering
\centering
\resizebox{0.2\textwidth}{!}{\begin{tabular}{l|l}
        \toprule
        Graph Type & Sufficiency \\
        \midrule
        Event Graph&92.0\\
        Scene Graph&21.0\\
        \bottomrule
    \end{tabular}}
    \caption{Comparison between the sufficiency of event graphs to scene graphs. Event graphs' high-level information allows them to be more expressive than scene graphs, thus more suitable for event-based tasks.}
    \label{table:event_graph_vs_scene_graph}
    \vspace{-0.6cm}
\end{table}

%% file: sec/5_conclusion.tex
\section{Conclusion}
This paper introduced \textbf{ENTER}, a novel interpretable VideoQA system that uses event graphs for contextually-aware and transparent video reasoning. ENTER transforms videos into event graphs allows it to bridge the gap between the interpretability of modular reasoning and the rich contextual awareness of bottom-up methods. Experiments show ENTER not only achieves state-of-the-art performance, but also offers a significantly more transparent and explainable reasoning process.

%% file: supp.tex
\clearpage
\setcounter{page}{1}
\appendix


\definecolor{backcolor}{rgb}{0.95,0.95,0.92}
\lstdefinestyle{mystyle}{
    backgroundcolor=\color{backcolor},   
    basicstyle=\ttfamily\scriptsize, 
    breaklines=true,                 
    frame=single,
    framesep=5pt,
    framerule=1pt,
}


\begin{figure*}[t]
    \centering
    \includegraphics[width=\linewidth]{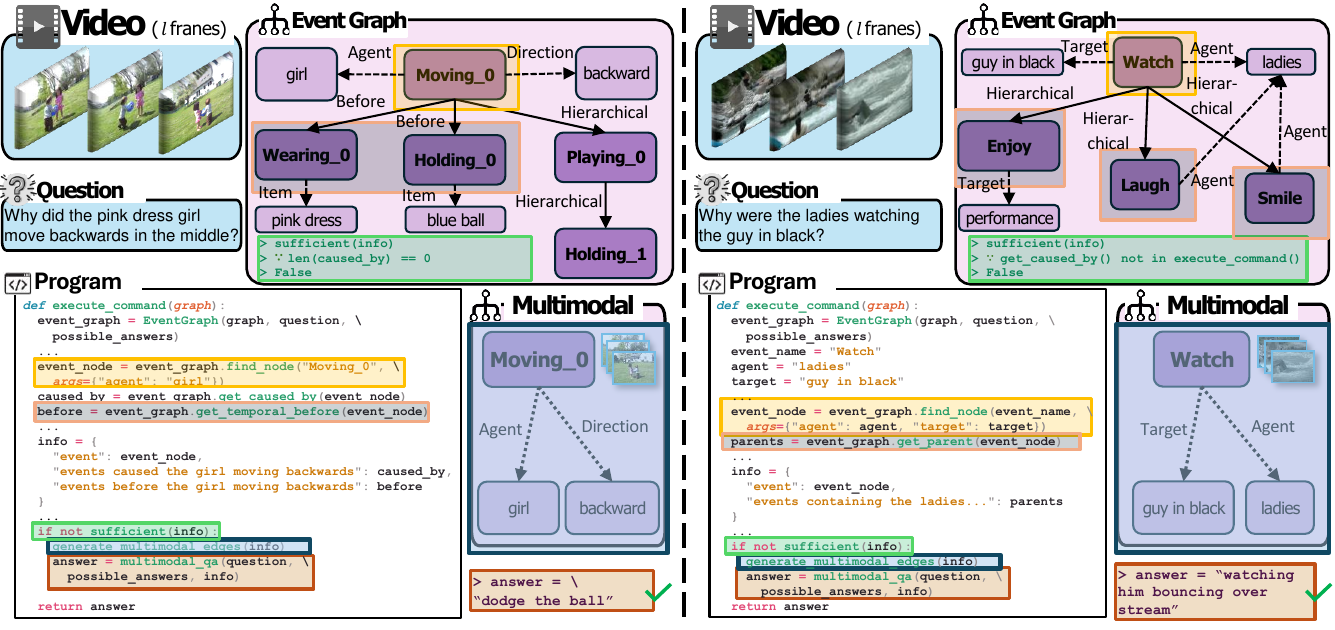}
    \caption{Qualitative examples of ENTER on NExT-QA, demonstrating interpretable reasoning alongside correct predictions.}
    \label{fig:samples_positive_supp}
\end{figure*}

\begin{figure*}[t]
    \centering
    \includegraphics[width=\linewidth]{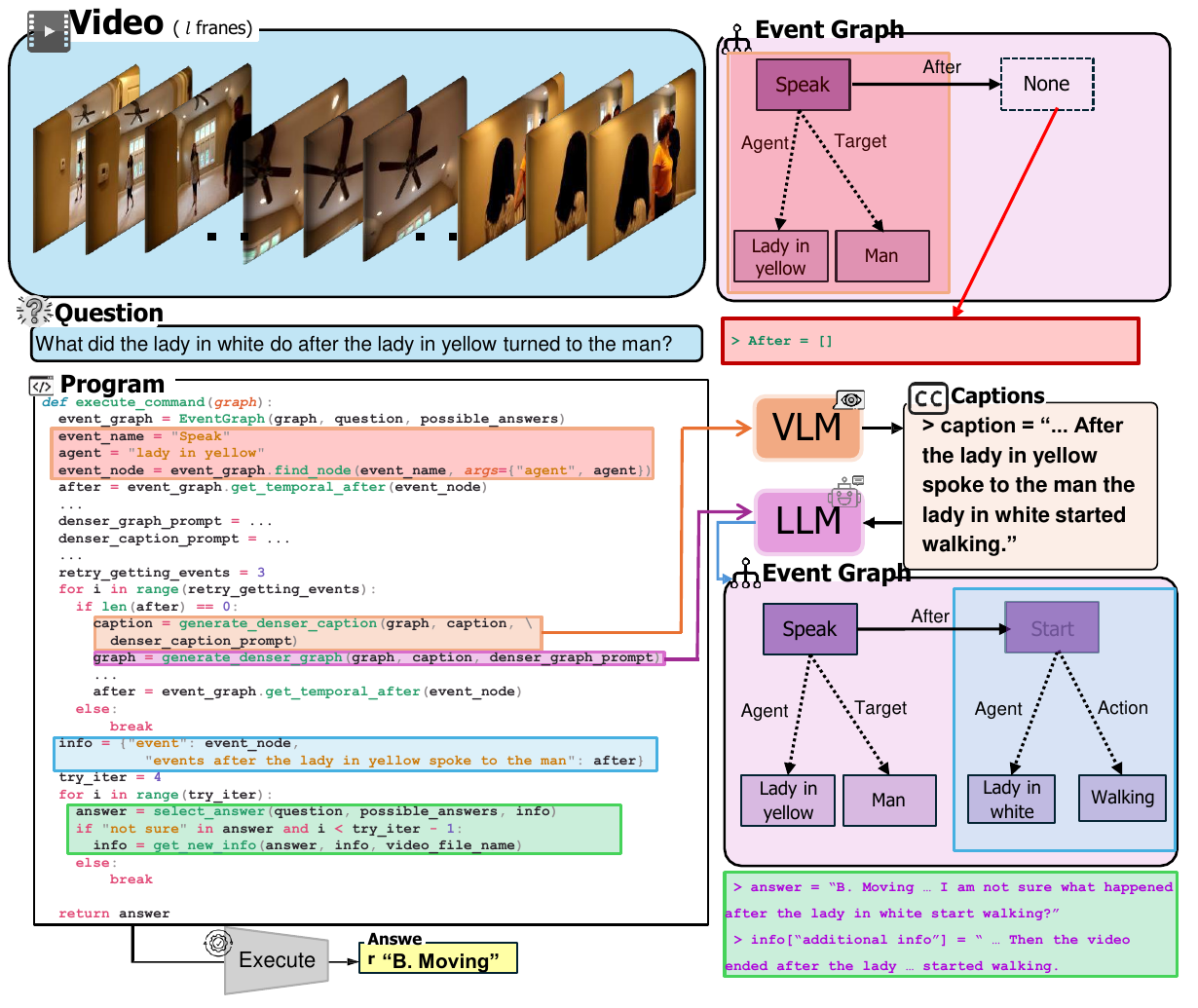}
    \caption{The figure demonstrates an example of case where \textit{generate\_denser\_caption} and \textit{generate\_denser\_graph} helps take care of failure case where text event graph is incomplete. The question asks about an event that occurred \textit{after} lady in yellow turned to the man. However, the initial graph didn't contain any after event. As such, \textit{generate\_denser\_caption} and \textit{generate\_denser\_graph} are called to include this missing information to finally arrive at the correct answer.}
    \label{fig:samples_positive_denser_gc}
\end{figure*}

We provide additional details in this section which further elucidate the claims and contributions of the paper. It's divided into the following sections:

\begin{itemize}
    \item Additional Baseline on STAR. (Appendix \ref{sec:additional_exp})
    \item Ablation on EgoSchema (Appendix \ref{sec:long_video_exp})
    \item Ablation on Hierarchical Update Mechanism (Appendix \ref{sec:hierarchical_update_ablation}) 
    \item More Qualitative Samples (Appendix \ref{sec:more_qualitative_samples})
    \item More Error Analysis (Appendix \ref{sec:error_analysis})
    \item Prompts (Appendix \ref{sec:prompts})\\
\end{itemize}

\section{Additional Baseline on STAR}
\label{sec:additional_exp}

Given that our method, ENTER, uses Gemini and GPT-4o as intermediate components, we also tested another modular method, TraveLER, by substituting its LLMs with Gemini and GPT-4o, for fair comparison. This \textit{upgraded} TraveLER baseline scores 65\% overall accuracy on STAR as compared to ENTER's 67.1\%. This further demonstrates that gains from ENTER are not merely a result of using superior LLMs, but rather originate from a superior methodology.

\section{Experiment on EgoSchema}
\label{sec:long_video_exp}
We perform an additional experiment on EgoSchema \cite{egoschema} to demonstrate our model ENTER’s robustness across diverse scenarios.
This dataset has long (180 seconds) egocentric videos with multiple-choice question-answering. 
It offers two evaluation splits: a hidden test set (Full set) of 5,000 videos evaluated via an external server, and a publicly available validation set (Subset) of 500 videos. We only consider the Full set in our evaluation. Although EgoSchema focuses on long video understanding rather than event analysis, we include it to show ENTER can be generalized to non-event-centric tasks.

\Cref{table:egoschema} compares our method to other modular methods on EgoSchema.
Our method surpasses all approaches on the EgoSchema dataset, with a 1.1\% improvement over ProViQ \citep{choudhury2023zeroshotvideoquestionanswering} and TraveLER \citep{shang2024travelermodularmultilmmagent}.
Although primarily designed for event-centric reasoning, our approach demonstrates stable performance on descriptive questions as well. 

\section{Ablation on Hierarchical Update Mechanism}
\label{sec:hierarchical_update_ablation}

To evaluate the contribution of each component described in \Cref{sec:method} to our method's performance, we conducted a detailed analysis of the percentage of data points activating each functionality and their corresponding accuracy. \Cref{table:multimodal_percent} presents an ablation study of the three hierarchical update mechanisms, detailing how frequently each component from \Cref{sec:method} is activated. The "Base" component refers to cases where the generated code operates without triggering additional modules like the "Denser Graph." Our results indicate that 63.6\% of data points achieve high accuracy without requiring additional visits to the video, demonstrating the baseline capability of our method to address a substantial portion of the dataset effectively. However, when critical information is overlooked during the initial video pass or cannot be adequately represented by unimodal graphs (as revealed in the qualitative error analysis), our method effectively identifies these scenarios and leverages the "Denser Graph," "Denser Caption," or "Multimodal Graph" functionalities. It is important to note that the subset activating "Multimodal Graph" overlaps with those activating "Denser Graph" and "Denser Caption," so the percentages do not sum to 100.

We also ran a qualitative experiment to judge the helpfulness of iterative graph update mechanism. We sample 30 graphs and ask human annotators to rate whether the information supplied in the update contains critical information to answer the question. The study reveals that when hierarchical update is called, 66.67\% time the update is helpful. Given that the answer accuracy of the samples where hierarchical update is called is ~71\% (Table \Cref{table:multimodal_percent}), being helpful 66.67\% time aligns well with the performance. These results prove and highlight the effectiveness of our hierarchical update mechanism

\input{tables/percent_multimodal}
\input{tables/egoschema}

\section{Qualitative Analysis}
\label{sec:more_qualitative_samples}

We further present two more positive examples in Figure \ref{fig:samples_positive_supp} to demostrate how the multimodal feature from our model and event graph can provide interpretability 

The first example (left figure) addresses a question about what caused a girl in a pink dress to move backward. Initially, the captioner and the model was able to construct an event graph and retrieved the event correctly, as shown in the light-orange highlighting. However, while the system identified the \textit{"Moving\textunderscore0"} event, it missed the crucial causal information explaining why the girl moved backward, as indicated by the green-highlighted code. To address this gap, the model employed the \textit{generate\_multimodal\_edges} function to analyze relevant clips and construct multimodal edges connected to the \textit{"Moving\_0"} event. This analysis revealed the missing causal action: \textit{"the girl dodging the ball,"} which led to the correct answer \textit{"dodge the ball."}

The second example (right figure) demonstrates a different scenario. The question references why the ladies were watching the guy in black; however, neither the event graph nor the generated code contains causal edges or code for retrieving the causal relationship for the event \textit{"Watch"}, making it insufficient to determine the answer. To resolve this, the model invokes the \texttt{generate\_multimodal\_edges} function to analyze relevant video clips for additional context. Through this analysis, it successfully identified the missing causal relationships and confirmed the final answer. These examples demonstrate ENTER's ability to incorporate visual information explicitly to recover from failure cases where information in the text-only event graph proves insufficient.

\Cref{fig:samples_positive_denser_gc} illustrates how \textit{generate\_denser\_caption} and \textit{generate\_denser\_graph} work to mitigate incomplete graph failure cases. The question in the figure asks about an event that occurred \textit{after} 'speaking' event. However, the initial event graph didn't contain any event with after edge from 'speaking' event. This triggered the \textit{generate\_denser\_caption} and \textit{generate\_denser\_graph} functions to try to incorporate this missing information. In this example, these two function successfully combined to include the missing information in the graph -- the fact that the lady started walking after 'speaking'. This led to the successful prediction of the correct answer.

\section{Error Analysis}
\label{sec:error_analysis}

\begin{figure*}[t]
    \centering
    \includegraphics[width=\linewidth]{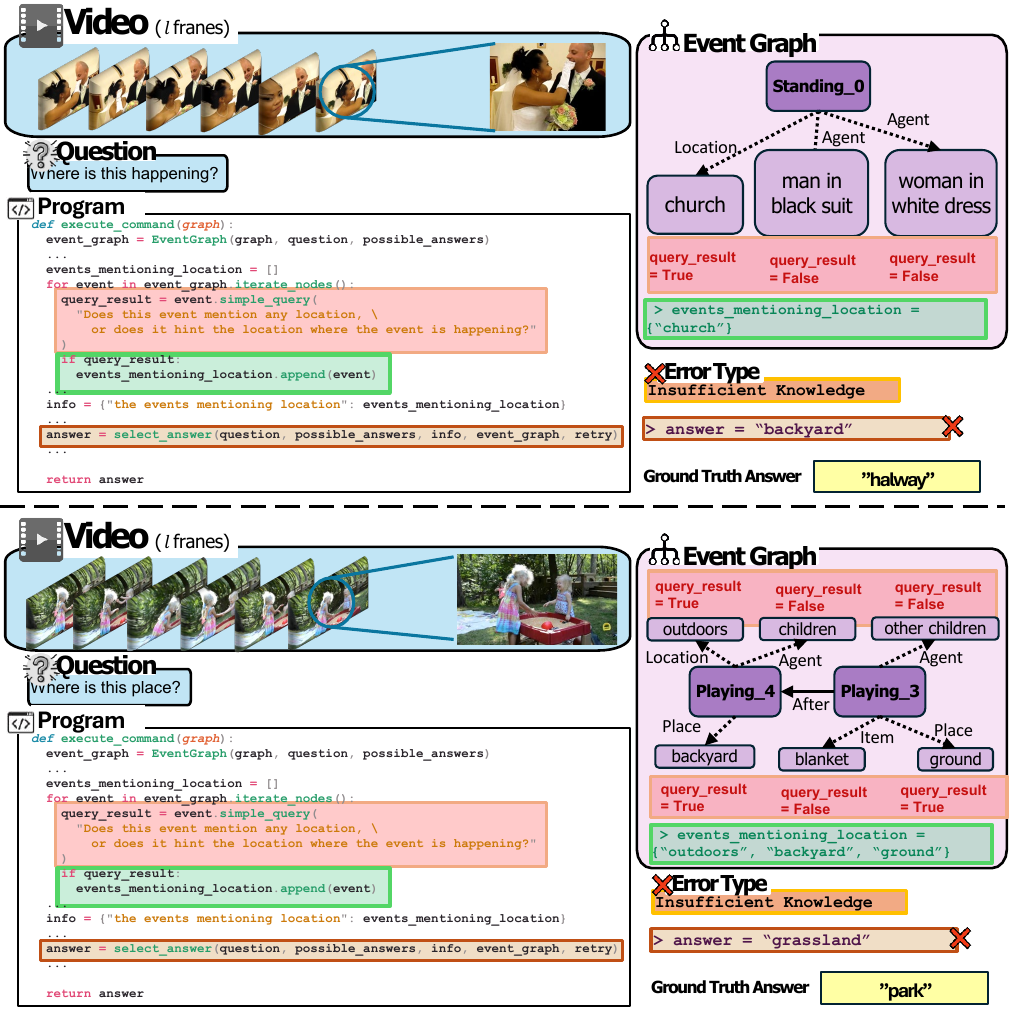}
    \caption{ENTER's interpretability analysis demonstrates two examples of insufficient knowledge errors (where information is not enough). The top example fails because the captioner failed to generate an event graph with the necessary location detail \textit{"hallway"} needed for the ground truth answer. Similarly, the bottom example fails because the event graph has no node containing information corresponding to the ground truth answer \textit{"park."}}
    \label{fig:error_supp_1}
\end{figure*}

\begin{figure*}[t]
    \centering
    \includegraphics[width=\linewidth]{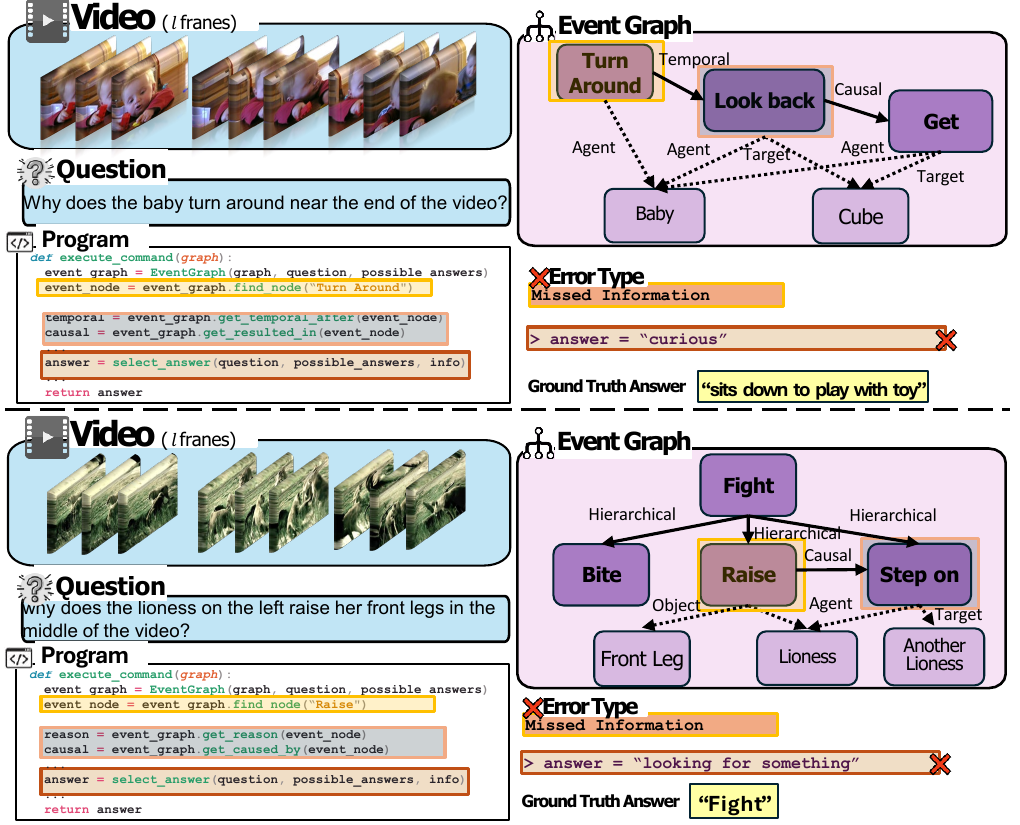}
    \caption{ENTER's interpretability analysis demonstrates two examples of missed information errors (where required information is missing). The top example fails because while the sequence of events \textit{"Turn Around"} to \textit{"Look back"} and \textit{"Look back"} to \textit{"Get"} are captured, the event graph misses the crucial causal edge from \textit{"Turn Around"} to events related to \textit{"Get"}, making the model unable to infer why the baby turned around. The bottom example shows a similar case where, although the basic events are captured, the event graph lacks the causal edge connecting \textit{"Raise"} to \textit{"Fight"}, preventing the model from understanding why the lion raised its front legs, leading to incorrect results.}
    \label{fig:error_supp_2}
\end{figure*}

\begin{figure*}[t]
    \centering
    \includegraphics[width=\linewidth]{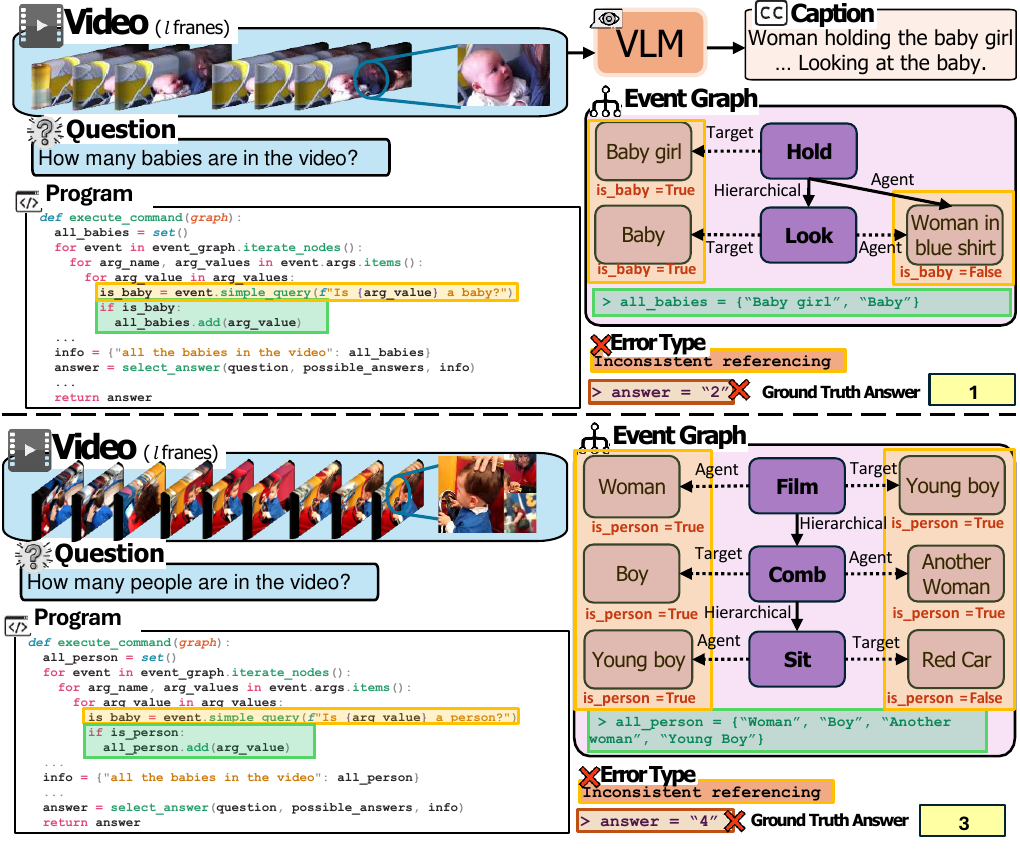}
    \caption{ENTER's interpretability analysis demonstrates two examples of inconsistent referencing errors (where the model misreferences objects or entities as the same or different). The top example shows the model incorrectly referencing the same baby from different frames as different objects, resulting in an incorrect result. Similarly, the bottom example shows incorrect referencing of the same group of people across different frames as a different group, leading to inaccurate counting of the number of people.}
    \label{fig:error_supp_3}
\end{figure*}

\input{figures/negative_examples}

To demonstrate our model's interpretability capabilities, we analyze examples of incorrect predictions to trace the source of different types of errors, including \textbf{Insufficient Knowledge}, \textbf{Missed Information} and \textbf{Inconsistent Referencing}.

\begin{itemize}[leftmargin=*]
\item\textbf{Insufficient Knowledge Leads to Incorrect Predictions Despite Correct Reasoning Steps} The insufficient knowledge error types occurs when the input lacks crucial information needed for accurate scene interpretation. 

\Cref{fig:error} top example highlights an error caused by the limited knowledge of the captioner. The question, “\textit{How do the two men play the instrument?}” should be answered with “\textit{Roll the handle.}” However, the captioner misidentifies the instrument as “\textit{look like a guitar,}”. The code fails to detect missing information in this case since it successfully retrieves the child node, even though with misinformation, resulting in the incorrect inference “\textit{strum the strings.}” The error shows how the captioner's misunderstanding of the instrument misleads the model's reasoning.

Figure \ref{fig:error_supp_1} illustrates two examples where the model attempts to identify video locations but fails due to insufficient knowledge. In the top example, although the model successfully generates the correct code to search for location information by traversing nodes in the event graph, it still fails to produce the correct answer due to missing critical information in the graph itself. While the ground truth location is \textit{"hallway"}, this information is absent from the captioner's output, resulting in an event graph that lacks any node containing the \textit{"hallway"} information. Instead, the event graph erroneously includes "church" as a location node. Thus, despite the model's successful retrieval of location-related nodes and correct code execution, it cannot arrive at the correct answer due to this fundamental gap in the input knowledge  Similarly, in the second example, the model retrieves location-related nodes including \textit{"outdoors"}, \textit{"backyard"} and \textit{"ground"}. However, none of these nodes correspond to the ground truth location. This again demonstrates how insufficient knowledge provided to the model can lead to prediction errors, even when the model's code functions properly.


\item\textbf{Missed Information Leads to Incorrect Event Reasoning} 

\Cref{fig:error_supp_2} presents two examples where the event graph captures basic events but fails to establish crucial causal relationships, leading to incorrect answers. In the top example, a child plays with toys and turns back at the end of the video to get a cube. While the sequence of events is captured in the graph with a temporal edge from \textit{"Turn Around"} to \textit{"Look back"} and a causal edge from \textit{"Look back"} to \textit{"Get"}, the event graph misses causal edges connecting the event \textit{"Turn Around"} to events related to the ground truth answer \textit{"Get"}. This missing causal relationship makes the model unable to infer the why the baby turned around, resulting in an incorrect prediction. Similarly, in the bottom example, a lion on the left raises its front legs in response to the lion on the right starting to fight. While these basic events are captured in the graph, the model cannot establish causal edges from the event \textit{"Raise"} to the event \textit{"Fight"}, as the event graph lacks information linking to the true cause, making the model unable to infer the actual reason why the lion raised her front legs.


\item\textbf{Inconsistent Referencing of Entities Result in Incorrect Answers}

\Cref{fig:error} shows the model missing the event “\textit{Trying to film}” which explains why the man is near the fighting field. Instead, the model guesses the wrong answer, “\textit{Wants to join the fight}” due to this oversight. The missed event leads the model to an incorrect inference about the man's intention.

\Cref{fig:error_supp_3} shows two examples where errors occur due to inconsistent entity referencing in the event graph. In the top example, the same baby is incorrectly referenced with two different labels: \textit{"Baby"} and \textit{"Baby girl"}. This inconsistency leads to the code double counting the number of babies in the video, resulting in an incorrect answer of two instead of one. Similarly, in the bottom example, the event graph generation shows the same referencing error. The video depicts a woman giving a haircut to a boy while another woman holds a camera, but the same child is referenced twice with different labels: \textit{"boy"} and \textit{"Young boy"}. Both examples illustrate how inconsistent entity referencing leads to counting errors in the final analysis.


\clearpage
\newpage
\onecolumn 
\section{Prompts}
\label{sec:prompts}

\begin{tcolorbox}[colback=backcolor, colframe=black!75!white, title=Captioner Prompt]
\begin{lstlisting}[style=mystyle]
You will be given a video. You need to describe the events in the video. Start your description with the first event happening in the video, describe all the events in chronological, causal, or hierarchical order, and end your description with the last event in the video. In your description, indicate temporal relations of events using keywords such as "after" "then" "meanwhile" etc.; indicate causal relations of events using key words such as "causing" etc.; indicate hierarchical relations of events using keywords such as "by" "with" etc.
When referring to entities, always use descriptive keywords representing their characteristics, such as "white Honda Civic", "man in black suit", "chair on the left", etc. Make sure the references to the entities are consistent, for example, if you mentioned a "black car with broken window", refer to it with the same name ("black car with broken window") everywhere else you mention it. For entities mentioned in the question, reference them using the same name everywhere else, for example, if the question mentions "man in black pants", use "man in black pants" to reference that entity.
Your description will be used to answer the following question: {question}? Choices: A. {a0} B. {a1} C. {a2} D. {a3} E. {a4}. When describing the video, do not answer the question, only describe the video in detail.
Example description: Two boxers, one in gold-white shorts, and the other one in orange-white shorts, facing each other in a boxing ring; meanwhile, the referee is observing the fight inside the ring; at the same time, a large number of audiences are watching. During the fight, the boxer in gold-white shorts throws a left jab, knocking down the opponent in orange-white shorts. After that, the referee counts to ten while the boxer in gold-white shorts raises his hand in victory; at the same time, the crowd cheers and applauds after the knockout. The referee announced the victory the boxer in gold-white victorious after he counted to ten; after that the boxer in gold-white shorts goes to a big screen and celebrates. Meanwhile, the referee and the boxer's cornermen check the well-being of the boxer in orange-white shorts.
\end{lstlisting}
\end{tcolorbox}

\begin{tcolorbox}[colback=backcolor, colframe=black!75!white, title=Graph Generator prompt - Part 1]
\begin{lstlisting}[style=mystyle]
Given a paragraph, please extract all events and its argument roles in the following format:
{<event_1>: {<argument_role_type_1>: [argument_role_1_1, argument_role_1_2, ...], <argument_role_type_2>: [argument_role_2_1, argument_role_2_2, ... ] , ....},
 <event_2>: {<argument_role_type_1>: [argument_role_1_1, argument_role_1_2, ...], <argument_role_type_2>: [argument_role_2_1,argument_role_2_2, ...], ...},
...}

Coreference of argument roles with different names should follow the name of the first occurrence of the argument. For example, \"blue car driving left\" and \"blue car turning left\" both should be named \"driving\". 

Coreference of events with different names should follow the name of the first occurrence of the argument. For example, \"blue car\" and \"blue vehicle\" both should be named \"blue car\". 

Collective arguments that are should maintain its finest grained form in the text. For example, if \"three cars\" containing \"red car\", \"blue car\", and \"green car\", then reference the three cars as [\"red car\", \"blue car\", \"green car\"] in the graph.

Differentiate events with the same name with indices. For example: \"man in blue shirt is eating pasta\" and \"woman is eating broccoli and rice\" should be \"{\"eating_0\": \"{\"agent\":[\"man in white\"], \"item\":[\"pasta\"]}, \"eating_1\": {\"agent\":[\"woman\"], \"item\":[\"rice\",\"broccoli\"]}}, this should also be reflected in the event graph.

In addition, extract all possible event-event relationships. The relationships should be from among causal, temporal and hierarchical relationships and the relationship should be directed.

The format should be:
{<event_1>: {causal: [<event_4>, <event_5>], temporal: [<event_2>], hierarchical: [<event_7>, <event_8>]},
<event_2>: {causal:[<event_1>]},
<event_3>:{},
....}

Causal Definition: An event A is causally related to event B if A was the cause of event B. For example: Black car hits the man in white. The man in white falls. Event relation: {\"Hit\": {\"causal\":[\"Fall\"]}}.

Temporal Definition: An event A is temporally related to event B if A started before B did.

\end{lstlisting}
\end{tcolorbox}

\begin{tcolorbox}[colback=backcolor, colframe=black!75!white, title=Graph Generator prompt - Part 2]
\begin{lstlisting}[style=mystyle]

Hierarchical Definition: An event A is hierarchically related to event B if B occurred spatiotemporally within A. Example: The man in white is fighting the person in monster costume. The man in white throws a punch. Event relation: {\"Fight\": {\"hierarchical\":[\"Throw\"]}}. Example: The waiter is serving the food by passing a tray. Event relation: {\"Serve\": {\"hierarchical\": [\"Pass\"]}}. Example: The soldier is holding his gun while patrolling. Event relation: {\"Patrol\": {\"hierarchical\": [\"Hold\"]}}.

In addition, you will also be given a question related to the text. You should focus on the events related to the question when generating the graph.

Example Text:
People are marching on a street during a protest. Police fired tear gas to disperse the crowd.

Example Question: Why did the police fired the tear gas?

Example Response:
Events:
{ 
\"Protest\": {\"agent\": [\"people\"], \"place\": \"street\", \"description\": \"People are protesting on the street.\"},
\"Marching\": {\"agent\": [\"people\"], \"place\": \"street\", \"description\": \"People are marching on the street.\"},
\"Fired\": {\"agent\": [\"police\"], \"item\": \"tear gas\". \"description\": \"Police fired tear gas.\"},
\"Disperse\": {\"item\": [\"crowd\"], \"description\": \"Crowd dispersed.\"}
}

Events-Events Relationships:
{
\"Protest\": {\"hierarchical\": [\"marching\", \"fired\", \"disperse\"]},
\"Marching\": {\"temporal\":[\"Fired\"]},
\"Fired\": {\"causal\":[\"disperse\"], \"temporal\":[\"disperse\"]},
\"Disperse\": {}
}

Example Question: Who came after the black car in the middle turned right?

Example Text:
Three cars are waiting at a traffic light. When it turned green, the red one went left, the black one on the right went right, and the middle black one went straight. After the black car turned right, a red tow truck drove through the road.

Example Response:
Events:
{ 
\"Waiting\": {\"agent\": [\"red car\", \"black car on the right\", \"middle black car\"], \"place\": [\"traffic light\"], \"description\": \"The red car, the black car on the right, and the black car in the middle are waiting for traffic light.\"},
\"Turned\": {\"agent\": [\"traffic light\"], \"color\": [\"green\"], \"description\": \"The traffic light turned green.\"},
\"Went_0\": {\"agent\": [\"red car\"], \"direction\": [\"left\"], \"description\": \"The red car turned left.\"},
\"Went_1\": {\"agent\": [\"black car on the right\"], \"direction\": [\"right\"], \"description\": \"The black car on the right turned right.\"},
\"Went_2\": {\"agent\": [\"middle black car\"], \"direction\": [\"straight\"], \"description\": \"The middle black car turned straight.\"},
\"Drive\": {\"agent\": [\"red tow truck\"], \"place\":[\"road\"], \"description\": \"The red tow truck drove through the road.\"}
}

\end{lstlisting}
\end{tcolorbox}

\begin{tcolorbox}[colback=backcolor, colframe=black!75!white, title=Graph Generator prompt - Part 3]
\begin{lstlisting}[style=mystyle]

Events-Events Relationships:
{
\"Waiting\": {\"temporal\": [\"Turned\"]},
\"Turned\": {\"causal\":[\"Went_0\", \"Went_1\", \"Went_2\", \"Drive\"], \"temporal\": [\"Went_0\", \"Went_1\", \"Went_2\"]},
\"Went_0\": {\"temporal\":[\"Drive\"]},
\"Went_1\": {\"temporal\":[\"Drive\"]},
\"Went_2\": {\"temporal\":[\"Drive\"]},
\"Drive\": {}
}

Question: What made the crowd applaud?

Example Text: 

Two boxers, one in gold-white shorts, and the other one in orange-white shorts, facing each other in a boxing ring. The boxer in gold-white shorts throws a left jab, knocking down the opponent in orange-white shorts. The referee counts to ten as the boxer in gold-white shorts raises his hand in victory. The crowd cheers and applauds. The boxer in gold-white shorts goes to a big screen and celebrates while the referee and the boxer's cornermen check the well-being of the boxer in orange-white shorts.

Example Response:

Events:
{
\"Facing\": {\"agent\": [\"boxer in gold-white shorts\", \"boxer in orange-white shorts\"], \"direction\":[\"each other\"], \"location\": [\"ring\"], \"description\": \"Boxer in gold-white shorts and boxer in orange-white shorts are facing each other in the ring.\"},
\"Throw\": {\"agent\": [\"boxer in gold-white shorts\"], \"action\": [\"left jab\"], \"description\":\"The boxer in gold-white shorts throws a left jab\"},
\"Knock down\": {\"agent\": [\"boxer in gold-white shorts\"], \"target\": [\"boxer in orange-white shorts\"], \"description\": \"The boxer in gold-white shorts knocks down the boxer in orange-white shorts\"},
\"Count to ten\": {\"agent\":[\"refree\"], \"description\": \"The refree counts to ten\"},
\"Raise\": {\"agent\": [\"boxer in gold-white shorts\"], \"item\":[\"hands\"], \"description\": \"Boxer in gold-white shorts raises his hands in victory.\"},
\"Applaud\": {\"agent\": [\"the crowd\"], \"description\": \"The crowd applauds.\"},
\"Go\": {\"agent\": [\"boxer in gold-white shorts\"], \"destination\":[\"big screen\"], \"description\": \"Boxer in gold-white shorts goes to the big screen\"},
\"Celebrate\": {\"agent\": [\"boxer in gold-white shorts\"], \"description\": \"Boxer in gold-white shorts celebrates.\"},
\"Check\": {\"agent\": [\"referee\", \"boxer's cornermen\"], \"target\":[\"boxer in orange-white shorts\"], \"description\": \"the referee and the boxer's cornermen check the well-being of the boxer in orange-white shorts.\"}
}

Events-Events Relationships:

{
\"Facing\": {\"temporal\": [\"Throw\"]},
\"Throw\": {\"causal\": [\"Knock down\"]},

\"Knock down\": {\"causal\": [\"Count to ten\", \"Raise\", \"Check\", \"Applaud\"], \"temporal\": [\"Count to ten\", \"Raise\", \"Applaud\"]},
\"Count to ten\": {\"temporal\":[\"Go\", \"Check\"]},
\"Raise\": {\"temporal\": [\"Go\", \"Check\"]},
\"Applaud\": {\"temporal\": [\"Go\", \"Check\"]},
\"Go\": {\"temporal\": [\"Celebrate\"]},
\"Celebrate\": [],
\"Check\": []
}
Given Text: 

\end{lstlisting}
\end{tcolorbox}

\begin{tcolorbox}[colback=backcolor, colframe=black!75!white, title=Code Generator Prompt - Part 1]
\begin{lstlisting}[style=mystyle]
Given the list of available python apis, please generate python code to answer the given question.

class Event:
    def __init__(self, name:str, args:dict):
        \"\"\"Initializes the event class with an event name and argument roles
        Parameters
        -------
        name: str
            name of the event
        args: dict
            event argument in this format: {arg_role: [arg_values]}
        \"\"\"

    def __str__(self,):
        return self.description

    def find_node(self, name, arguments):
        \"\"\" Returns an event node from the event graph given the event name and the associated arguments
        -------
        >>> # Find the event "riding" with agent "man in hat"
        >>> def execute_command(graph, caption, question, possible_answers, video_file_name):
        >>>     event_graph = EventGraph(graph, question, possible_answers)
        >>>     event_name = \"riding\"
        >>>     agent = \"man in hat\"
        >>>     event_node = event_graph.find_node(event_name, args={\"agent\": agent})
        \"\"\"

    def simple_query(self, query):
        \"\"\"Returns the answer to a basic question yes-or-no asked about arguments of the event, returns either True or False. The questions are about basic information about the event, and are not meant to be used for complex reasoning for the event, reasoning between among events, or external knowledge.
        Examples
        -------
        >>> # Find if the event "running" has a dog in it
        >>> def execute_command(graph, caption, question, possible_answers, video_file_name):
        >>>     event_graph = EventGraph(graph, question, possible_answers)
        >>>     event_name = \"running\"
        >>>     event_node = event_graph.find_node(event_name)
        >>>     has_dog = event_node.simple_query("Is there a dog in this event?.") # Ask an simple yes-or-no question to determine if the event contains a dog
        >>>     if has_dog:
        >>>         return 'yes'
        >>>     else:
        >>>         return 'no'
        \"\"\"

class EventGraph:
    def __init__(self, event_graph, question, possible_answers):
        \"\"\"Initializes an event graph. Also creates an Event item for each event.
        Each event is stored internally as a string but interfaces with outside class with Event items.
        Parameters
        -------
        event_graph: Tuple[dict,dict]
            event_graph in the following format: ({ <event_1>: {<argument_role_type_1>: argument_role_1, <argument_role_type_2>: argument_role_2, ....},
                                     <event_2>: {<argument_role_type_1>: argument_role_1, <argument_role_type_2>: argument_role_2, ....},
                                    ...}, # Containing events and their arguments,
            {<event_1>: {causal: [<event_4>, <event_5>], temporal: [<event_2>], hierarchical: [<event_7>, <event_8>]},
                            <event_2>: {causal:[<event_1>]},
                            <event_3>:{},
                            ....} # Containing relationships between events)

        \"\"\"
        self.event_graph, self.events = self.parse_events(event_graph)
        self.event_graph = event_graph
        self.events_dict = {}
        for event, args in events_json.items():
            self.events_dict[event] = Event(event, args)

\end{lstlisting}
\end{tcolorbox}

\begin{tcolorbox}[colback=backcolor, colframe=black!75!white, title=Code Generator Prompt - Part 2]
\begin{lstlisting}[style=mystyle]

    def iterate_nodes(self):
        \"\"\" Iterates through all nodes in the graph, to be used in a for loop
        >>> # How many people are there in this video? 
        >>> def execute_command(graph, caption, question, possible_answers, video_file_name):
        >>>     event_graph = EventGraph(graph, question, possible_answers)
        >>>     all_people = set()
        >>>     for event in event_graph.iterate_nodes():
        >>>         for arg_name, arg_values in event.args.items():
        >>>             for arg_value in arg_values: # arg_values: List[str] contains all the arguments under arg_name
        >>>                 is_person = event.simple_query(f\"Is {arg_value} a person?\") # Ask an simple yes-or-no question to determine if the argument is a person or not
        >>>                 if is_person:
        >>>                     all_people.add(arg_value)
        >>>     info = {\"all the people in the video\": all_people}
        >>>     answer = select_answer(question, possible_answers, info, event_graph)
        >>>     return answer, info, event_graph\"\"\"
        return find_node(event_graph, node)

    def find_node(self, event_name, args=None):
        \"\"\"Returns an event node from the event graph given the event name and the associated arguments
        >>> # Find the event "knight riding horse"
        >>> def execute_command(graph, caption, question, possible_answers, video_file_name):
        >>>     event_graph = EventGraph(graph, question, possible_answers)
        >>>     event_name = \"riding\"
        >>>     agent = \"knight\"
        >>>     item = \"horse\"
        >>>     event_node = event_graph.find_node(event_name, args={\"agent\": agent, \"item\": item})
        >>>     info = {\"event\": event_node}
        >>>     answer = select_answer(question, possible_answers, info, event_graph)
        >>>     return answer, info, event_graph\"\"\"
        return find_node(event_graph, node)

    def get_children(self, node):
        \"\"\"Returns all the children nodes of the specified event node
        >>> # Find the child events of "how are the protesters getting more attention"
        >>> def execute_command(graph, caption, question, possible_answers, video_file_name):
        >>>     event_graph = EventGraph(graph, question, possible_answers)
        >>>     event_name = \"getting attention\"
        >>>     agent = \"protesters\"
        >>>     event_node = event_graph.find_node(event_name, args={\"agent\": agent})
        >>>     children = event_graph.get_children(event_node)
        >>>     info = {\"event\": event_node, \"events contained by the protesters getting attention\": children}
        >>>     answer = select_answer(question, possible_answers, info, event_graph)
        >>>     return answer, info, event_graph
        \"\"\"
        return get_children(event_graph, node)

    def get_parent(self, node):
        \"\"\"Returns the parent node of the specified event node
        >>> # Find the parent events of "why is the man cutting vegetables"
        >>> def execute_command(graph, caption, question, possible_answers, video_file_name):
        >>>     event_graph = EventGraph(graph, question, possible_answers)
        >>>     event_name = \"Cut\"
        >>>     agent = \"man\"
        >>>     item = \"vegetables\"
        >>>     event_node = event_graph.find_node(event_name, args={\"agent\": agent, \"item\": item})
        >>>     parent = event_graph.get_parent(event_node)
        >>>     caused_by = event_graph.get_caused_by(event_node)
        >>>     info = {\"event\": event_node, \"events containing man cutting vegetables\": parent, \"events caused man cutting vegetables\": caused_by}
        >>>     answer = select_answer(question, possible_answers, info, event_graph)
        >>>     return answer, info, event_graph
        \"\"\"
        
        return get_parent(event_graph, node)
\end{lstlisting}
\end{tcolorbox}

\begin{tcolorbox}[colback=backcolor, colframe=black!75!white, title=Code Generator Prompt - Part 3]
\begin{lstlisting}[style=mystyle]

    def get_temporal_after(self, node):
        \"\"\"Returns all the events that happened after the specified event node
        >>> # Find the events happened after "man in blue driving black car"
        >>> def execute_command(graph, caption, question, possible_answers, video_file_name):
        >>>     event_graph = EventGraph(graph, question, possible_answers)
        >>>     event_name = \"driving\"
        >>>     agent = \"man in blue\"
        >>>     item = \"black car\"
        >>>     event_node = event_graph.find_node(event_name, args={\"agent\": agent, \"item\": item})
        >>>     after = event_graph.get_temporal_after(event_node)
        >>>     info = {\"event\": event_node, \"events after man in blue driving black car\": after}
        >>>     answer = select_answer(question, possible_answers, info, event_graph)
        >>>     return answer, info, event_graph
        \"\"\"
        return temporal_after(node)
        
    def get_temporal_before(self, node):
        \"\"\"Returns all the events that happened before the specified event node
        
        >>> # Find the events happened before "the white horse with number 3 passed the finish line"
        >>> def execute_command(graph, caption, question, possible_answers, video_file_name):
        >>>     event_graph = EventGraph(graph, question, possible_answers)
        >>>     event_name = \"pass\"
        >>>     agent = \"white horse with number 3\"
        >>>     location = \"finish line\"
        >>>     event_node = event_graph.find_node(event_name, args={\"agent\": agent, \"location\": location})
        >>>     before = event_graph.get_temporal_before(event_node)
        >>>     info = {\"event\": event_node, \"the white horse with number 3 passed the finish line\": before}
        >>>     answer = select_answer(question, possible_answers, info, event_graph)
        >>>     return answer, info, event_graph
        \"\"\"
        return temporal_before(node)

    def get_resulted_in(self, node):
        \"\"\"Returns the event that resulted in specified event node
        >>> # Find the events causing "the soldier camouflage fell to the ground"
        >>> def execute_command(graph, caption, question, possible_answers, video_file_name):
        >>>     event_graph = EventGraph(graph, question, possible_answers)
        >>>     event_name = \"fall\"
        >>>     agent = \"soldier in camouflage\"
        >>>     event_node = event_graph.find_node(event_name, args={\"agent\": agent})
        >>>     resulted_in = event_graph.get_resulted_in(event_node) 
        >>>     before_event = event_graph.get_temporal_before(event_node) 
        >>>     info = {\"event\": event_node, \"The soldier camouflage fell to the ground because of these events\": resulted_in, \"These events happened before the soldier in camouflage fell to the ground\": before_event}
        >>>     answer = select_answer(question, possible_answers, info, event_graph)
        >>>     return answer, info, event_graph
        \"\"\"
        return caused_by(node)
        
    def get_caused_by(self, node):
        \"\"\"Returns all the events that happened due to the specified event node"
        >>> # Find the events happened due to "the samurai in Japanese armor wielding his katana"
        >>> def execute_command(graph, caption, question, possible_answers, video_file_name):
        >>>     event_graph = EventGraph(graph, question, possible_answers)
        >>>     event_name = \"wield\"
        >>>     agent = \"samurai in Japanese armor\"
        >>>     item = \"katana\"
        >>>     event_node = event_graph.find_node(event_name, args={\"agent\": agent, \"item\": item})
        >>>     happened_after = event_graph.get_temporal_after(event_node)
        >>>     caused_by = event_graph.get_caused_by(event_node)
        >>>     info = {\"event\": event_node, \"The samurai in Japanese armor wielding his sword caused these events to happen\": caused_by, \"These events happened after the samurai in Japanese armor wielding his sword\": happened_after}
        >>>     answer = select_answer(question, possible_answers, info, event_graph)
        >>>     return answer, info, event_graph
        \"\"\"
        return resulted_in(node)

\end{lstlisting}
\end{tcolorbox}

\begin{tcolorbox}[colback=backcolor, colframe=black!75!white, title=Code Generator Prompt - Part 4]
\begin{lstlisting}[style=mystyle]
def select_answer(question, possible_answers, info, event_graph):
    \"\"\"Selects an Answer from the possible answers given the question and the info in a dictionary format.\"\"\"
    return select_answer(question, possible_answers, info, event_graph)

Question: How many cats are there in the video? A. one B. two C. three D. four E. five

def execute_command(graph, caption, question, possible_answers, video_file_name):
    event_graph = EventGraph(graph, question, possible_answers)
    events_mentioning_location = []
    
    all_cats = set()
    for event in event_graph.iterate_nodes():
        for arg_name, arg_values in event.args.items():
            for arg_value in arg_values: # arg_value: List[str] contains the arguments under arg_name
                is_cat = event.simple_query(f\"Is {arg_value} a cat?\")
                if is_cat:
                    all_cats.add(arg_value)
    info = {\"all the cats in the video\": all_cats}
    answer = select_answer(question, possible_answers, info, event_graph)
    return answer

Example Question: Where is the video taking place? A. Road B. House C. Dog D. Dining room E. street

def execute_command(graph, caption, question, possible_answers, video_file_name):
    event_graph = EventGraph(graph, question, possible_answers)
    events_mentioning_location = []
    for event in event_graph.iterate_nodes():
        query_result = event.simple_query(\"Does this event mention any location, or does it hint the location where the event is happening?\")
        if query_result:
            events_mentioning_location.append(event)
    info = {\"the events mentioning location\": events_mentioning_location}
    retry_answering = 3
    answer = select_answer(question, possible_answers, info, event_graph)
    return answer

Example Question: how is the man in blue feeling after standing up from the chair? A. Happy B. Sad C. Angry D. Neutral E. Surprised

def execute_command(graph, caption, question, possible_answers, video_file_name):
    event_graph = EventGraph(graph, question, possible_answers)
    event_name = \"stand up\"
    agent = \"man in blue\"
    item = \"chair\"
    event_node = event_graph.find_node(event_name, args={\"agent\": agent, \"item\": item})
    
    after = event_graph.get_temporal_after(event_node)
    children = event_graph.get_children(event_node)
    caused_by = event_graph.get_caused_by(event_node)

\end{lstlisting}
\end{tcolorbox}

\begin{tcolorbox}[colback=backcolor, colframe=black!75!white, title=Code Generator Prompt - Part 5]
\begin{lstlisting}[style=mystyle]
    retry_getting_events = 3
    denser_graph_prompt = \"\"\"Read the question and the paragraph, and find relations and/or events in the paragraph that are related to the event the man in blue standing up. Use these events to generate a new graph, such that the new graph has at least one of the following for question answering: the event after \"stand up\", what happens after the man in blue standing up, if you find these events, create these new events and draw the event relation \"stand up\": {\"temporal\":[\"<event>\", ...]} for these events; the event contained by \"stand up\", what is the man doing while he is standing up, if you find these events, create these new events and draw the event relation \"stand up\": {\"hierarchical\":[<event>, ...]} for these events; the event caused by \"stand up\", what happened due to the man standing up, if you find these events, create these new events and draw the event relation \"stand up\": {\"causal\":[<event>, ...]} for these events.\"\"\"
    denser_caption_prompt = \"\"\"Watch the video, describe what happens after the man in blue standing up; is he doing while standing up; what event happened due to the man standing up; and everything else that is helpful in answering the question.\"\"\"
    if len(after)+len(children)+len(caused_by) == 0:
        graph = generate_denser_graph(event_node, graph, caption, question, possible_answers, denser_graph_prompt, 0)
        event_graph = EventGraph(graph, question, possible_answers)
        event_node = event_graph.find_node(event_name, args={\"agent\": agent, \"item\": item})
        after = event_graph.get_temporal_after(event_node)
        children = event_graph.get_children(event_node)
        caused_by = event_graph.get_caused_by(event_node)
    for i in range(retry_getting_events):
        if len(after)+len(children)+len(caused_by) == 0:
            caption = generate_denser_caption(event_node, graph, caption, question, possible_answers, denser_caption_prompt, video_file_name, i)
            graph = generate_denser_graph(event_node, graph, caption, question, possible_answers, denser_graph_prompt, i+1)
            event_graph = EventGraph(graph, question, possible_answers)
            event_node = event_graph.find_node(event_name, args={\"agent\": agent, \"item\": item})
            after = event_graph.get_temporal_after(event_node)
            children = event_graph.get_children(event_node)
            caused_by = event_graph.get_caused_by(event_node)
        else:
            break
    info = {\"event\": event_node,
            \"events after the man in blue standing up\": after,
            \"events containing the man in blue standing up\": children,
            \"events caused by the man in blue standing up\": caused_by
            }
    retry_answering = 3
    for i in range(retry_answering):
        answer = select_answer(question, possible_answers, info, event_graph)
        if \"not sure\" in answer and i < retry_answering - 1:
            info = get_new_info(answer, info, graph, caption)
        else:
            break
    return answer

Example Question: whom was the police firing the tear gas to? A. Robbers B. Terrorists C. Protesters D. Attackers E. Prisoners

\end{lstlisting}
\end{tcolorbox}

\begin{tcolorbox}[colback=backcolor, colframe=black!75!white, title=Code Generator Prompt - Part 6]
\begin{lstlisting}[style=mystyle]

Python Code:
def execute_command(graph, caption, question, possible_answers, video_file_name):        
    event_graph = EventGraph(graph, question, possible_answers)
    event_name = \"fire\"
    agent = \"police\"
    item = \"tear gas\"
    event_node = event_graph.find_node(event_name, args={\"agent\": agent, \"item\": item})

    resulted_in = event_graph.get_resulted_in(event_node)
    parents = event_graph.get_parent(event_node)
    caused_by = event_graph.get_caused_by(event_node)
    
    retry_getting_events = 3
    denser_graph_prompt = \"\"\"Read the question and the paragraph, and find relations and/or events in the paragraph that are related to the event the police firing tear gas. Use these events to generate a new graph, such that the new graph has at least one of the following for answering the question: the event causing \"fire\", what events directly caused the police to fire tear gas, if you find these events, create these new events and draw the event relation <event>: {\"causal\":[\"fire\", ...]} for these events; the event contains \"fire\", what do they achieve by firing the tear gas, if you find these events, create these new events and draw the event relation <event>: {\"hierarchical\":[\"fire\", ...]} for these events; the event caused by \"fire\", what happened due to the firing of tear gas, if you find these events, create these new events and draw the event relation \"fire\": {\"causal\":[<event>, ...]} for these events.\"\"\"
    denser_caption_prompt = \"\"\"Watch the video, describe what causes the police to fire the tear gas; what do they need to achieve to fire the tear gas; what event happened due to the police firing tear gas; and everything else that is helpful in answering the question.\"\"\"
    if len(caused_by)+len(parents)+len(resulted_in) == 0:
        graph = generate_denser_graph(event_node, graph, caption, question, possible_answers, denser_graph_prompt, 0)
        event_graph = EventGraph(graph, question, possible_answers)
        event_node = event_graph.find_node(event_name, args={\"agent\": agent, \"item\": item})
        resulted_in = event_graph.get_resulted_in(event_node)
        parents = event_graph.get_parent(event_node)
        caused_by = event_graph.get_caused_by(event_node)
    for i in range(retry_getting_events):
        if len(caused_by)+len(parents)+len(resulted_in) == 0:
            caption = generate_denser_caption(event_node, graph, caption, question, possible_answers, denser_caption_prompt, video_file_name, i)
            graph = generate_denser_graph(event_node, graph, caption, question, possible_answers, denser_graph_prompt, i+1)
            event_graph = EventGraph(graph, question, possible_answers)
            event_node = event_graph.find_node(event_name, args={\"agent\": agent, \"item\": item})
            resulted_in = event_graph.get_resulted_in(event_node)
            parents = event_graph.get_parent(event_node)
            caused_by = event_graph.get_caused_by(event_node)
        else:
            break

    info = {\"event\": event_node,
            \"events resulted in police firing tear gas\": resulted_in,
            \"events containing the police firing tear gas\": parents,
            \"events caused by police firing tear gas\": caused_by
            }

    retry_answering = 3
    for i in range(retry_answering):
        answer = select_answer(question, possible_answers, info, event_graph)
        if \"not sure\" in answer and i < retry_answering - 1:
            info = get_new_info(answer, info, graph, caption)
        else:
            break
    return answer, info, event_graph

\end{lstlisting}
\end{tcolorbox}
\begin{tcolorbox}[colback=backcolor, colframe=black!75!white, title=Code Generator Prompt - Part 7]
\begin{lstlisting}[style=mystyle]

Example Question: how did the man in blue shirt cook the steak? A. Hold with hands B. Put on a grill C. Fry in pan D. Boil in water E. Putting it in the oven
Python Code:
def execute_command(graph, caption, question, possible_answers, video_file_name):
    event_graph = EventGraph(graph, caption, video_file_name)
    event_name = \"cook\"
    agent = \"man in blue shirt\"
    item = \"steak\"
    # Get the event node related to the question
    event_node = event_graph.find_node(event_name, args={\"agent\": agent, \"item\": item})

    children = event_graph.get_children(event_node)
    before = event_graph.get_temporal_before(event_node)

    retry_getting_events = 3
    denser_graph_prompt = \"\"\"Read the question and paragraph, and find relations and/or events that are related to the man is cooking steak in the paragraph. Use these events to generate a new graph, such that the new graph has one of the following for question answering: the events contained by \"cook\", focus on what tools techniques and methods is he using, if you find these events, create new event nodes for them and add the event relation \"cook\":{\"hierarchical\":[<event>, ...]} for these events; the events before the man cooking the steak, focus on the possible preparation he did for the cooking, if you find these events, create new events for them and add the event relation <event>:{\"temporal\":[\"cook\", ...]}. \"\"\"
    denser_caption_prompt = \"\"\"Watch the video, describe: what happens before the man in blue shirt cooking the steak; what is the environment is it like; what is he doing while cooking the steak; what method, technique or tool he is using to cook the steak; and everything else that is helpful in answering the question.\"\"\"
    if len(children)+len(before) == 0: 
        graph = generate_denser_graph(event_node, graph, caption, question, possible_answers, denser_graph_prompt, 0)
        event_graph = EventGraph(graph, question, possible_answers)
        event_node = event_graph.find_node(event_name, args={\"agent\": agent, \"item\": item})
        children = event_graph.get_children(event_node)
        before = event_graph.get_temporal_before(event_node)
    for i in range(retry_getting_events):
        if len(children)+len(before) == 0:
            caption = generate_denser_caption(event_node, graph, caption, question, possible_answers, denser_caption_prompt, video_file_name, i)
            graph = generate_denser_graph(event_node, graph, caption, question, possible_answers, denser_graph_prompt, i+1)
            event_graph = EventGraph(graph, question, possible_answers)
            event_node = event_graph.find_node(event_name, args={\"agent\": agent, \"item\": item})
            children = event_graph.get_children(event_node)
            before = event_graph.get_temporal_before(event_node)
        else:
            break

    info = {\"event\": event_node, \"events contained by the man cooking steak\": children, \"events before the man cooking steak\": before}
    
    retry_answering = 3
    for i in range(retry_answering):
        answer = select_answer(question, possible_answers, info, event_graph)
        if \"not sure\" in answer and i < retry_answering - 1:
            info = get_new_info(answer, info, graph, caption)
        else:
            break
    return answer, info, event_graph

\end{lstlisting}
\end{tcolorbox}
\begin{tcolorbox}[colback=backcolor, colframe=black!75!white, title=Code Generator Prompt - Part 8]
\begin{lstlisting}[style=mystyle]

Example Question: how does the kid get the ball at the beginning? A. With his hand B. Kicks it C. Run to fetch it D. Looks at the baby E. Happy
def execute_command(graph, caption, question, possible_answers, video_file_name):
    event_graph = EventGraph(graph, question, possible_answers)
    event_name = \"get\"
    agent = \"kid\"
    item = \"ball\"
    # Get the event node related to the question
    event_node = event_graph.find_node(event_name, args={\"agent\": agent, \"item\": item})

    children = event_graph.get_children(event_node)
    before = event_graph.get_temporal_before(event_node)

    retry_getting_events = 3
    denser_graph_prompt = \"\"\"Read the question and paragraph, and find relations and/or events that are related to the kid getting the ball in the paragraph. Use these events to generate a new graph, such that the new graph has one of the following for question answering: the events contained by \"get\", focus on what actions he is performing in order to get the ball, if you find these events, create new event nodes for them and add the event relation \"get\":{\"hierarchical\":[<event>, ...]} for these events; the events before the kid getting the ball, focus where is the ball located, what action the kid is doing before getting the ball, if you find these events, create new events for them and add the event relation <event>:{\"temporal\":[\"get\", ...]}. \"\"\"
    denser_caption_prompt = \"\"\"Watch the video, describe: what happens before the kid getting the the ball; where is the ball located; what is the kid doing while getting the ball; what action is the kid doing to get the ball; and everything else that is helpful in answering the question.\"\"\"
    if len(children) + len(before) == 0:
        graph = generate_denser_graph(event_node, graph, caption, question, possible_answers, denser_graph_prompt, 0)
        event_graph = EventGraph(graph, question, possible_answers)
        event_node = event_graph.find_node(event_name, args={\"agent\": agent})
        children = event_graph.get_children(event_node)
        before = event_graph.get_temporal_before(event_node)
    for i in range(retry_getting_events):
        if len(children) + len(before) == 0:
            caption = generate_denser_caption(event_node, graph, caption, question, possible_answers, denser_caption_prompt, video_file_name, i)
            graph = generate_denser_graph(event_node, graph, caption, question, possible_answers, denser_graph_prompt, i+1)
            event_graph = EventGraph(graph, question, possible_answers)
            event_node = event_graph.find_node(event_name, args={\"agent\": agent})
            children = event_graph.get_children(event_node)
            before = event_graph.get_temporal_before(event_node)
        else:
            break

    info = {\"event\": event_node, 
            \"child events of the kid getting the ball\": children,
            \"before the kid getting the ball\": before}
    
    retry_answering = 3
    for i in range(retry_answering):
        answer = select_answer(question, possible_answers, info, event_graph)
        if \"not sure\" in answer and i < retry_answering - 1:
            info = get_new_info(answer, info, graph, caption)
        else:
            break
    return answer, info, event_graph

Example Question: what did the hen do after the chicks wandered off? A. Moves the food in her hand B. Eat the corn C. Lay down on the straw D. Flip onto back E. look around
\end{lstlisting}
\end{tcolorbox}
\begin{tcolorbox}[colback=backcolor, colframe=black!75!white, title=Code Generator Prompt - Part 9]
\begin{lstlisting}[style=mystyle]
Python Code: 
def execute_command(graph, caption, question, possible_answers, video_file_name):
    event_graph = EventGraph(graph, question, possible_answers)
    event_name = \"wander\"
    agent = \"chicks\"
    # Get the event node related to the question
    event_node = event_graph.find_node(event_name, args={\"agent\": agent})

    after = event_graph.get_temporal_after(event_node)
    caused_by = event_graph.get_caused_by(event_node)

    retry_getting_events = 3
    denser_graph_prompt = \"\"\"Read the question and paragraph, and find relations and/or events that are related to the chicks wandered off that describe what did the hen do in the paragraph. Use these events to generate a new graph such that the new graph has one of the following for question answering: the events after \"wander\", focus on what was the behavior of the hen, or what did it interact with, if you find these events, create new event nodes for them and add the event relation \"wander\":{\"temporal\":[<event>, ...]} for these events; the events caused by \"wander\", focus on what caused the hen did due to the chicks wandered off, what was its reaction to the chicks, if you find these events, create new events for them and add the event relation \"wander\":{\"causal\":[<event>, ...]}. \"\"\"
    denser_caption_prompt = \"\"\"Watch the video, describe: what happened to the hen after the chicks wandered off; what was the hen's behavior after the chicks wandered off; the chicks wandered off caused the hen to do what; and everything else that is helpful in answering the question.\"\"\"
    if len(after)+len(caused_by) == 0: 
        graph = generate_denser_graph(event_node, graph, caption, question, possible_answers, denser_graph_prompt, 0)
        event_graph = EventGraph(graph, question, possible_answers)
        event_node = event_graph.find_node(event_name, args={\"agent\": agent})
        after = event_graph.get_temporal_after(event_node)
        caused_by = event_graph.get_caused_by(event_node)
    for i in range(retry_getting_events):
        if len(after)+len(caused_by) == 0: 
            caption = generate_denser_caption(event_node, graph, caption, question, possible_answers, denser_caption_prompt, video_file_name, i)
            graph = generate_denser_graph(event_node, graph, caption, question, possible_answers, denser_graph_prompt, i+1)
            event_graph = EventGraph(graph, question, possible_answers)
            event_node = event_graph.find_node(event_name, args={\"agent\": agent})
            after = event_graph.get_temporal_after(event_node)
            caused_by = event_graph.get_caused_by(event_node)
        else:
            break

    info = {\"event\": event_node, \"events after the chicks wandered off\": after, \"events caused by the chicks wandered off\": caused_by}
    
    retry_answering = 3
    for i in range(retry_answering):
        answer = select_answer(question, possible_answers, info, event_graph)
        if \"not sure\" in answer and i < retry_answering - 1:
            info = get_new_info(answer, info, graph, caption)
        else:
            break
    return answer, info, event_graph    

For the provided question and graph below, please generate python code to solve it. When you are looking for events by name, make sure to specify the exact name, include the indicies if they are present in the graph (i.e.Run_0, Eat_1). Only generate the execute_command method, do not generate anything else. 
\end{lstlisting}
\end{tcolorbox}

\begin{tcolorbox}[colback=backcolor, colframe=black!75!white, title=Denser Graph Prompt]
\begin{lstlisting}[style=mystyle]
    input_text = f"{graph_prompt}\n Paragraph: {caption} \n Graph: {original_graph} \n You need to gennerate a new graph based on the original graph that has additional events and/or relations that addresses the following concern: {request}. The new graph will be used to answer the following question Question: {question}? {choices} \n Response: \n Events: \n" where graph prompt is the prompt for graph generator
\end{lstlisting}
\end{tcolorbox}

\begin{tcolorbox}[colback=backcolor, colframe=black!75!white, title=Denser Caption Prompt]
\begin{lstlisting}[style=mystyle]
You will be given a video and a request. The request is made to ask for specific details the provided description is missing. You need to generate a new description that addresses the requested details. \n In your description, indicate temporal, causal, or hierarchical relations of events clearly using keywords such as \"at the same time\", \"mean while\", \"after\", \"causing\", \"due to\", etc.\n When referring to entities, always use descriptive keywords representing their characteristics, such as \"white Honda Civic\", \"woman in purple dress\", \"man in black suit\", \"chair on the left and chair on the right\", etc. Make sure the references to the entities are consistent, for example, if you mentioned a \"black car with broken window\", refer to it with the same name (\"black car with broken window\") everywhere else you mention it. When making the description, do not answer the question.\n Now, generate a new text addressing the concerns stated in the request: {request}:
\end{lstlisting}
\end{tcolorbox}

\begin{tcolorbox}[colback=backcolor, colframe=black!75!white, title= Event Simple Query Prompt]
\begin{lstlisting}[style=mystyle]
[
        {"role": "system", "content": "You will be given a description to an event and a simple question. Answer the question with \"yes\" or \"no\" based on the information provided. Do not generate anything else besides \"yes\" or \"no\"."},
        {"role": "user", "content": f"Event: {self.description}.\n Question: {query}. Answer the question with \"yes\" or \"no\" only.\n Your answer:\n"}
]
\end{lstlisting}
\end{tcolorbox}

\begin{tcolorbox}[colback=backcolor, colframe=black!75!white, title= Reasoner Prompt]
\begin{lstlisting}[style=mystyle]
[{"role": "system",
             "content": "You will be given a set of information describing events in a video and a multiple choice question. You need to answer the question with the information from the paragraph. Chose one and only one of the answer from the five choices by returning the corresponding letter from A-E. You must choose one as your final answer, or make an educated guess. If the information provided is insufficient to answer the question, you must: First, guess an answer by choosing one of the choices from A, B, C, D, or E. Then, indicate you are not sure by saying \"I am not sure\". Finally, explain what additional information you need, and explain what details or events you will focus on to obtain the information if you are watching the video. Only say you are not sure when the events do not mention what is being asked."},
            {"role":"user",
             "content":[
                {"type":"text",
                  "text": user_input}
]}]
\end{lstlisting}
\end{tcolorbox}

\begin{tcolorbox}[colback=backcolor, colframe=black!75!white, title= Get Additional Info Prompt]
\begin{lstlisting}[style=mystyle]
You will be given a video, a question, and a textual request generated by a language model asking for additional information to answer the question. You should watch the video, read the question and the request, then generate a textual description of the video focusing on what is being asked in the request. You should not try to determine if it answered correctly, nor answer the question directly. You should generate information addressing what the request is asking for, additionally you can also generate information helpful for answering the question. \n Question: {question} \n Choices: {choices} \n Request: {concern}. \n Your response: '
\end{lstlisting}
\end{tcolorbox}

\begin{tcolorbox}[colback=backcolor, colframe=black!75!white, title= Reasoning with multimodal graph prompt]
\begin{lstlisting}[style=mystyle]
You will be given a set of multimodal information as a multimodal graph, a question, and a set of choices. You need to answer the question with the multimodal graph by selecting from one of the choices, represented by one letter. Your answer must contain a single letter only without any additional text. \n Question: {question} Choices: {choices}
\end{lstlisting}
\end{tcolorbox}

\end{itemize}

%% file: tables/percent_multimodal.tex
\begin{table} [t]\small
    \centering
   
    \fontsize{8}{9}\selectfont
    \setlength{\tabcolsep}{3pt}
    \begin{tabular}{l|l|l}
        \toprule
        Activated Component& Percent & Subset Accuracy\\
        \midrule
        Base&63.6&78.6\\
        Denser Graph&15.3&71.9\\
        Denser Caption&4.8&81.9\\
        Multimodal Graph&20.9&62.5\\
        \bottomrule
    \end{tabular}
     \caption{Subset sizes of samples in NextQA that activate different hierarchical update mechanisms. The majority of the answers are sufficiently answerable by the base method without activating any additional mechanisms.}
    \label{table:multimodal_percent}
\end{table}

%% file: tables/egoschema.tex
\begin{table} [t]\
    \centering
    
\resizebox{0.9\linewidth}{!}{
    \begin{tabular}{l|c}
        \toprule
        Method & EgoSchema Fullset\\
        \midrule
        ProViQ \citep{choudhury2023zeroshotvideoquestionanswering}&53.3\\
        MoReVQA \citep{morevqa}&51.7\\
        TraveLER \citep{shang2024travelermodularmultilmmagent}&53.3\\
        ViperGPT \citep{vipergpt}&-\\
        Ours & \textbf{54.4}\\
        \bottomrule
    \end{tabular}
    }
    \caption{Our method EgoSchema compared with other zero-shot methods and modular methods.}
    \label{table:egoschema}
    \vspace{-0.3cm}
\end{table}

%% file: figures/negative_examples.tex
\begin{figure*}[t]
    \vspace{-1cm}
    \includegraphics[width=\textwidth,keepaspectratio]{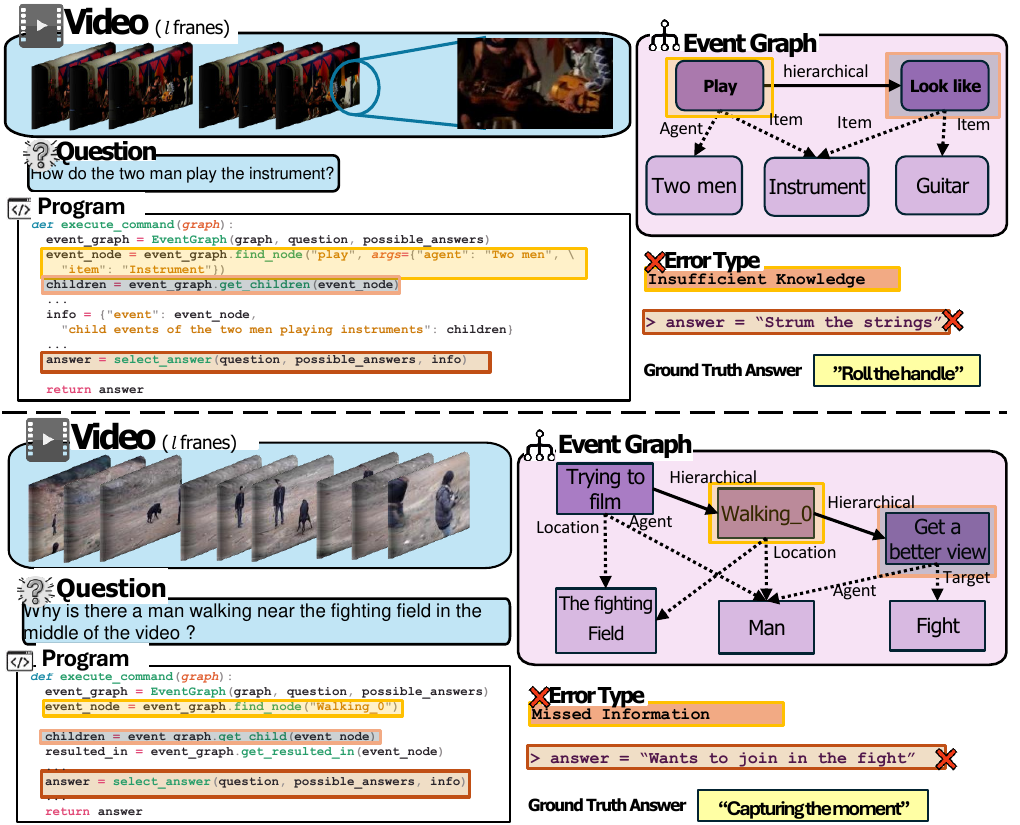}
    \caption{
    ENTER's interpretability allows the diagnosis and classification of errors into three major types: insufficient knowledge, inconsistent referencing, and missed information. Here, we depict insufficient knowledge and missing information error type.}
    \label{fig:error}
    \vspace{-0.8cm}
\end{figure*}